\documentclass{bmvc2k}

\usepackage{amssymb}
\usepackage{soul}
\usepackage{xcolor}
\usepackage[most]{tcolorbox}
\usepackage{array}
\usepackage{booktabs}
\usepackage{makecell} % For line breaks in cells
\usepackage{multirow}
\usepackage{multicol}

\title{CRCE: Coreference-Retention Concept Erasure in Text-to-Image Diffusion Models}

\addauthor{Yuyang Xue}{yuyang.xue@ed.ac.uk}{1}
\addauthor{Edward Moroshko}{emoroshk@ed.ac.uk}{1}
\addauthor{Feng Chen}{feng.chen@ed.ac.uk}{1}
\addauthor{Jingyu Sun}{s2091784@ed.ac.uk}{1}
\addauthor{Steven McDonagh}{smcdonag@ed.ac.uk}{1}
\addauthor{Sotirios A. Tsaftaris}{s.tsaftaris@ed.ac.uk}{1}

\addinstitution{
School of Engineering\\
University of Edinburgh\\
Edinburgh, UK
}

\runninghead{Student, Prof, Collaborator}{CRCE: Coreference-Retention Concept Erasure}

\def\eg{\emph{e.g}\bmvaOneDot}

\begin{document}

\maketitle

\begin{abstract}
Text-to-Image diffusion models can produce undesirable content that necessitates concept erasure. However, existing methods struggle with under-erasure, leaving residual traces of targeted concepts, or over-erasure, mistakenly eliminating unrelated but visually similar concepts. 
To address these limitations, we introduce \emph{CRCE}, a novel concept erasure framework that leverages Large Language Models to identify both semantically related concepts that should be erased alongside the target and distinct concepts that should be preserved. 
By explicitly modelling coreferential and retained concepts semantically, \emph{CRCE} enables more precise concept removal, without unintended erasure. 
Experiments demonstrate that \emph{CRCE} outperforms existing methods on diverse erasure tasks, including real-world object, person identities, and abstract intellectual property characteristics. 
The constructed dataset \emph{CorefConcept} and the source code will be release upon acceptance.
\end{abstract}

\begin{figure}[!thb]
  \centering
  \includegraphics[width=\linewidth]{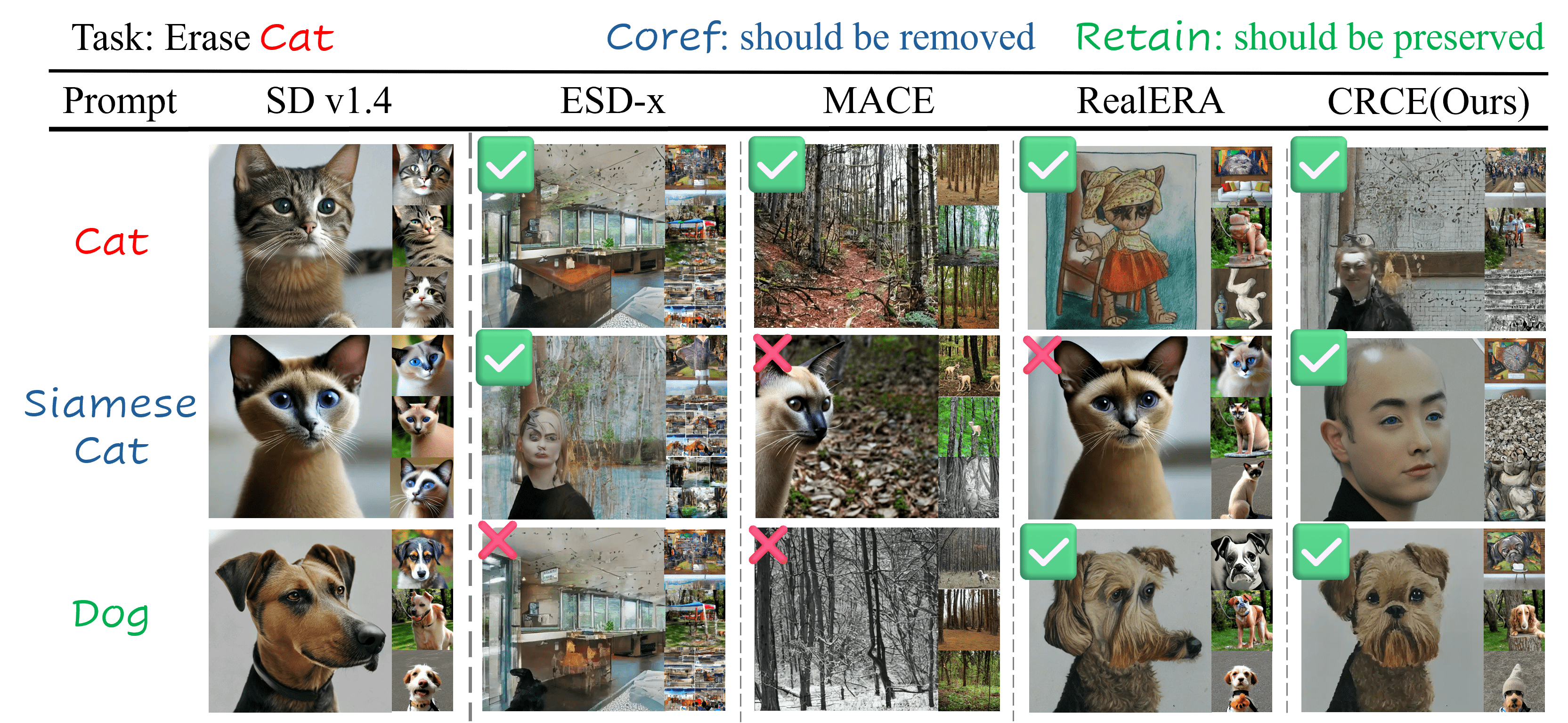}
  \caption{Consider the task of erasing the ``{Cat}'' concept. We define ``Siamese cat'' as coreference (coref), which should \emph{also} be erased, and ``Dog'' as a retain concept, \textit{i.e.} which should \emph{not} be erased. We show examples from SD v1.4 before any erasure and proceed with results from concept erasure methods. Green checkmarks (\color{green}{\checkmark}\color{bmv@captioncolor}) indicate successful erasure or retentions, while red crosses (\color{red}$\times$\color{bmv@captioncolor}) highlight failures. Our approach effectively balances erasure and retention, reducing both under- and over-erasure issues compared to existing methods.}
  \label{fig:1}
\end{figure}

%-------------------------------------------------------------------------
\section{Introduction}
\label{sec:intro}
% Introduce T2I model and its potential risks
Text-to-Image (T2I) diffusion models have demonstrated a remarkable ability to generate diverse, high-fidelity images from text prompts \cite{rombach2022high,saharia2022photorealistic,ramesh2022hierarchical}.
However, these powerful generative models may also produce undesirable concepts gleaned from extensive training corpora such as LAION-5B~\cite{schuhmann2022laion}.
These concepts encompass \textit{pornographic content}, \textit{violent} or \textit{hateful imagery}~\cite{zhang2024generate}, \textit{copyrighted art styles} or \textit{specific private identities}~\cite{jiang2023ai,somepalli2023diffusion}, \textit{social biases}~\cite{luccioni2023stable,struppek2023exploiting}, and other \textit{sensitive} materials~\cite{qu2023unsafe,schramowski2023safe}.
Although the public release of these T2I models has made image generation technology widely accessible, the presence of such undesirable concepts raises serious legal, safety, and ethical concerns~\cite{roose2022ai,hunter2023ai,Setty2023ai}. Consequently, efforts have emerged to remove specific concepts so that models can no longer reproduce them -- an approach collectively referred to as \textbf{concept erasure}.

% Introduce concept erasing and its goal
Concept erasure modifies a pre-trained generative model to eliminate a target concept from its output while preserving the model's ability of generating the remaining concepts~\cite{kumari2023ablating,zhang2024forget}.
Unlike post-intervention approaches such as filtering outputs and employing safety checkers at runtime~\cite{rombach2022high}, concept erasure pursues proactive solutions that alter the model's parameters, effectively removing the concept such that it cannot be easily evaded by cunning prompts Red Teaming~\cite{tsai2023ring} or direct weight access~\cite{gandikota2023erasing}. 

% The observation of under/over concept erasure.
However, despite progress in the field, we corroborate recent findings~\cite{liu2024realera,chowdhury2025fundamental} that identify persistent issues, even with recent models. Fig.~\ref{fig:1} shows this graphically. Erasing the concept ``cat'' can affect an unrelated concept ``dog'', while semantically related concepts such as ``Siamese Cat'' may resist erasure. During concept erasure, some \textbf{coreferential} concepts (\textit{corefs}) may remain \textit{under-erased}, (or have \textit{concept residues}~\cite{liu2024realera}), whereas \textbf{retained}, unrelated, concepts (\textit{retains}) may be removed (\textit{over-erased}).
These issues occur not only with objects but also with identities. Although methods based on Euclidean distance, cosine similarity, and CLIP~\cite{radford2021learning} have shown progress (further discussed in Sec.~\ref{sec:relatedwork}), these challenges remain largely unresolved. 

In this paper, we propose an intuitive yet effective concept erasure framework named Coreference-Retention Concept Erasure (\textit{CRCE}). Given a target concept, we employ a large language model (LLM) to generate semantically accurate \textit{coref} and \textit{retain} concepts.  
These corefs and retains are jointly integrated in a new loss function, which effectively balances the erasure of a target concept (and its corefs) whilst preserving unrelated knowledge. \emph{CRCE} can be easily generalised to diffusion-based models due to its flexibility and simplicity.
%\yuyang{The loss can be generalised to any diffusion-based models for its flexibility and simplicity.}

%we can still add another line or two

Our \textbf{main contributions} are summarised below:
\begin{enumerate}
    \item To the best of our knowledge, \textit{CRCE} %is 
    represents the first method to simultaneously address both under- and over-erasure of concepts by leveraging LLMs to identify semantic relationships between concepts along their natural manifolds rather than relying on neither cosine similarity nor Euclidean proximity.
    \item We propose and validate a novel loss function that jointly integrates
    LLM-generated coref and retain concepts with weighted certainty levels,
    enabling targeting of concept manifolds while preserving unrelated knowledge.
    Comprehensive experiments
    %and studies
    show that our method outperforms existing concept erasure techniques.
    %\yuyang{Comprehensive experiments and studies show our method outperforming existing erasure techniques.}
    \item We introduce \emph{CorefConcept}, a comprehensive dataset that maps semantic relationships between target concepts and their corefs/retains with quantified certainty metrics, providing a valuable resource for future research on concept erasure and understanding of semantic relationships in embedding spaces.
\end{enumerate}

\section{Preliminaries}
\label{sec:Preliminaries}
To set the scene for our approach, we start by introducing useful preliminary knowledge about the background and limitations of existing concept erasure methods, particularly regarding assumptions on the embedding space of CLIP.
%\yuyang{\st{Here, we introduce foundational components and methods on which our work builds on and gets inspired from. We begin with an overview of LDM, which serve as the backbone of T2I diffusion models, followed by a discussion of current concept erasure methods that form the baseline for our approach. We then discuss limitations in current methods regarding their understanding of semantic relationships in the embedding space to motivate our solution.}}

% \begin{figure}[t]
%   \centering
%   \includegraphics[width=0.8\linewidth]{images/fig2-min.png}
%    \caption{
%    Under-/Over-Erasure issues for the ESD method.
%    }
%    \label{fig:2}
% \end{figure}

\subsection{Baseline: Erasing Stable Diffusion (ESD)}
\label{sec:ESD}
 ESD~\cite{gandikota2023erasing} is a self-distillation method; a frozen copy of the original model guides a fine-tuned copy to eliminate the concept.
Consider a pre-trained Stable Diffusion (SD) model with weights $\theta^*$,
a target concept embedding $c$ that requires removal, and a copy of the model with weights $\theta$ that needs to be fine-tuned.
%\yuyang{needs to be fine-tuned}. 
During fine-tuning, ESD generates synthetic training noise by using the frozen model $\theta^*$ to guide the tuned model away from $c$. 
At each training step, a random noise latent $x_t$ is generated at a specific diffusion timestep $t$, which is input into the tuned model $\theta$, conditioned on a prompt containing concept $c$. 
Concurrently, the same noise is input into the frozen model $\theta^*$ with both the concept $c$ and an unconditional empty prompt.
%``''. 
These inputs are used to calculate an anchor noise prediction that represents the intended target concept.
The tuned model $\theta$ is updated to align predictions more closely with this anchor. 
The process employs classifier-free guidance, described by:
\begin{equation*}
  \epsilon_{\theta^*}^{\text{anchor}}(x_t, c, t) = \epsilon_{\theta^*}(x_t,t)-\eta[\epsilon_{\theta^*}(x_t, c, t) - \epsilon_{\theta^*}(x_t, t)],
  %\label{eq:ESD_diffusion}
\end{equation*}
where $\eta$ denotes the strength of the negative guidance. The fine-tuning objective seeks to make the 
%conditional predictions 
conform to this guided anchor, %following:
by minimizing\footnote{We consistently omit $\mathbb{E}_{z_t, t, c, \epsilon\sim\mathcal{N}(0,1)}$ for the sake of simplicity. %for all loss functions.%\edward{is it really needed?}
}:
\begin{equation}
  \mathcal{L}_{\text{ESD}}=
  %\mathbb{E}_{z_t, t, c, \epsilon}\left[
  \|\epsilon_\theta(x_t, c, t)-\epsilon_{\theta^*}^{\text{anchor}}(x_t, c, t)\|^2_2~.
  %\right]
  \label{eq:ESD_diffusion}
\end{equation}
Importantly, ESD's
%\yuyang{'s \st{does not require actual images of $c$; its}} 
beauty lies in utilizing the frozen model to instruct the newly adjusted model only using inherent knowledge, simply reversing the concept's influence.
However, ESD has a key limitation: it focuses on removing the specific target concept without considering semantically related or unrelated concepts, as the  example in Fig.~\ref{fig:1} illustrates.
%\yuyang{as an example illustrated in Fig.~\ref{fig:1}.\st{, when erasing the concept ``cat", semantically related concepts such as ``Siamese cat" often remain and can still be generated, resulting in under-erasure. Conversely, when erasing a concept such as ``deer", unrelated but visually similar concepts such as ``antelope" or ``gazelle" may be inadvertently removed, demonstrating over-erasure. While the ESD authors acknowledge this issue in their appendix, %by testing with paraphrased prompts, the method itself lacks any systematic mechanism to address the conceptual relationships between the target and its semantic neighbours.}} 
Hence methods that better account for the semantic relationships between concepts are needed.

\subsection{Related Work on Addressing Under-/Over-Erasure}
\label{sec:relatedwork}
%\yuyang{
Numerous methods aim to balance concept removal and utility in T2I models but do not address over-/under-erasure in totality. 
Techniques like CA~\cite{kumari2023ablating} ablating the anchor to its super-class (\eg~``grumpy cat'' to ``cat'') and attention-resteering FMN~\cite{zhang2024forget} suffer from under-erasure or unanalysed trade-offs. Closed-form editing methods such as UCE~\cite{gandikota2024unified} and MACE~\cite{lu2024mace} overlook synonym leakage or rely on non-automated semantic discovery. Others like SPM~\cite{lyu2024one}, 
%EUC~\cite{bui2024erasing}, 
RECE~\cite{gong2024reliable}, Receler~\cite{huang2024receler}, and AdvUnlearn~\cite{zhang2024defensive} use heuristics, shifting retains, red-teaming, gradient cues, or fixed banks, respectively, without fully resolving both erasure issues simultaneously. Retain set selection
%, as in Bui~\etal
\cite{bui2025fantastic} can be problematic for abstract concepts and super-classes. While all above methods use anchors, regularisers, or adversarial retains to protect unrelated content, none jointly optimise a certainty-weighted set of coreferential and retain prompts, a gap our method addresses.
A more comprehensive review of related work on latent diffusion models and concept erasure is provided in Appendix~\ref{sec:app_related_work}.

\subsection{CLIP Embedding Similarity vs. Semantic Similarity}
\label{sec:clip}

\begin{figure}[t]
  \centering
  \includegraphics[width=0.8\linewidth]{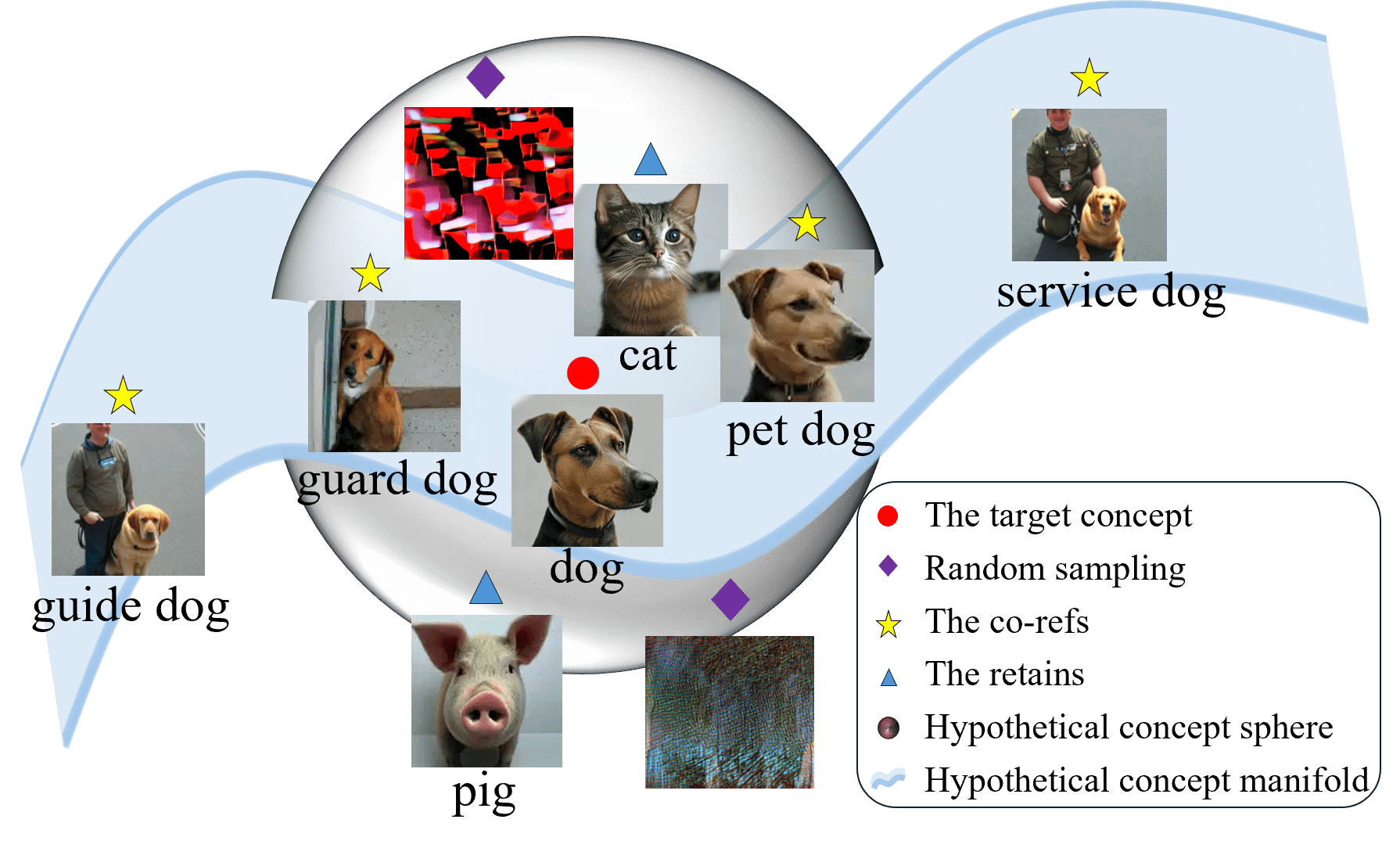}
   \caption{
    Illustration of how related and unrelated concepts to “dog” are arranged in CLIP's embedding space. The red dot marks the target concept ``dog”; yellow stars (\eg ``guide dog”, ``service dog”) represent coreferent concepts along the same semantic manifold, while blue triangles (\eg ``cat”, ``pig”) denote unrelated concepts to be retained. Neglecting corefs and retains leads to under-/over-erasure. RealEra~\cite{liu2024realera} samples random corefs in a spherical region, which poorly approximates the true semantic geometry—often capturing unrelated concepts (purple diamonds) - and ignores the non-Euclidean nature of concept relationships, where semantically distinct concepts may appear close in Euclidean space (blue triangles).
   %This figure illustrates 
   %Illustration of how related and unrelated concepts to ``dog'' cluster in CLIP's embedding space. 
   %This affects ESD~\cite{gandikota2023erasing} and suggests that the approximation of a sphere by RealEra~\cite{liu2024realera} is inadequate. }
   %The central red dot represents the target concept ``dog'', while yellow stars indicate coref concepts (\eg ``guide dog'' and ``service dog'') that lie on the same hypothetical concept manifold.
   %Blue triangles represent concepts that should be retained (``cat'' and ``pig'') that exist outside this manifold. 
   %Considering only the target concept but not the corefs and retains, causing under-/over-erasure issues.
   %RealEra~\cite{liu2024realera} uses random corefs sampling in a spherical region around the target concept, which fails to capture the true geometry of concept distribution, and has the potential to capture irrelevant or meaningless concepts (purple diamonds).
   %It fails to account for the non-Euclidean nature of semantic relationships, where non-related concepts like ``cat" may be deceptively close in Euclidean distance to ``dog" despite being semantically distinct. 
   %The placement of each point in the schematic is representative of calculated Euclidean distance scale. 
   %at the same scale of the calculated Euclidean distance.
   }
   \label{fig:3}
\end{figure}

Trained on a massive dataset of image-text pairs, CLIP~\cite{radford2021learning} encodes images and text into a shared embedding space using a contrastive objective, making it a popular component in T2I models. 
CLIP similarity is a popular metric for exploiting and evaluating text/image embeddings by measuring the cosine similarity between their normalized vectors.
%\yuyang{\st{or by the Euclidean distance}}. 
Ideally, cosine similarity within CLIP's embedding space would reflect human notions of semantic similarity—conceptually similar images should be positioned close together, while dissimilar ones should be spread further apart. 
However, recent studies have demonstrated that human perception of semantic similarity does not always correspond with cosine distances within CLIP's representation space~\cite{fu2023dreamsim,li2024erroneous}. 
In certain cases, CLIP similarity can appear counterintuitive from a human perspective, revealing a discrepancy between the model's learnt representations and human semantics. Fig.~\ref{fig:3} illustrates the concept of dispersed prompt embeddings, in CLIP space, that are not clustered around the concept centre.
% That means that building on clip similarity for unlearning might be a problem...
When examining Euclidean distances, we find counterintuitive relationships: ``dog''-``cat'' (0.419) and ``dog''-``pig'' (0.466) are actually closer in the CLIP embedding space than ``dog''-``service dog'' (0.533) and ``dog''-``guide dog'' (0.539). 
%\yuyang{\st{This contradicts human semantic understanding, where ``service dogs'' and ``guide dogs'' are types of dogs, yet they appear more distant than entirely different animal species.}} 
Appendix~\ref{sec:clipdistance} provides an extensive list of distances. 

We argue that making inferences on the geometry of the CLIP embedding space (as RealEra~\cite{liu2024realera} does) is not the appropriate approach to fully determine how a model should erase concepts. Our approach instead bypasses this issue by explicitly finding sets of concepts that should be used as corefs and those that shall remain (retain) to directly guide the erasure process to reduce instances of under- and over-erasure.

\section{Method: Coref-Retain Concept Erasure (CRCE)}
\label{sec:method}

\begin{figure*}[t]
  \centering
  \includegraphics[width=\linewidth]{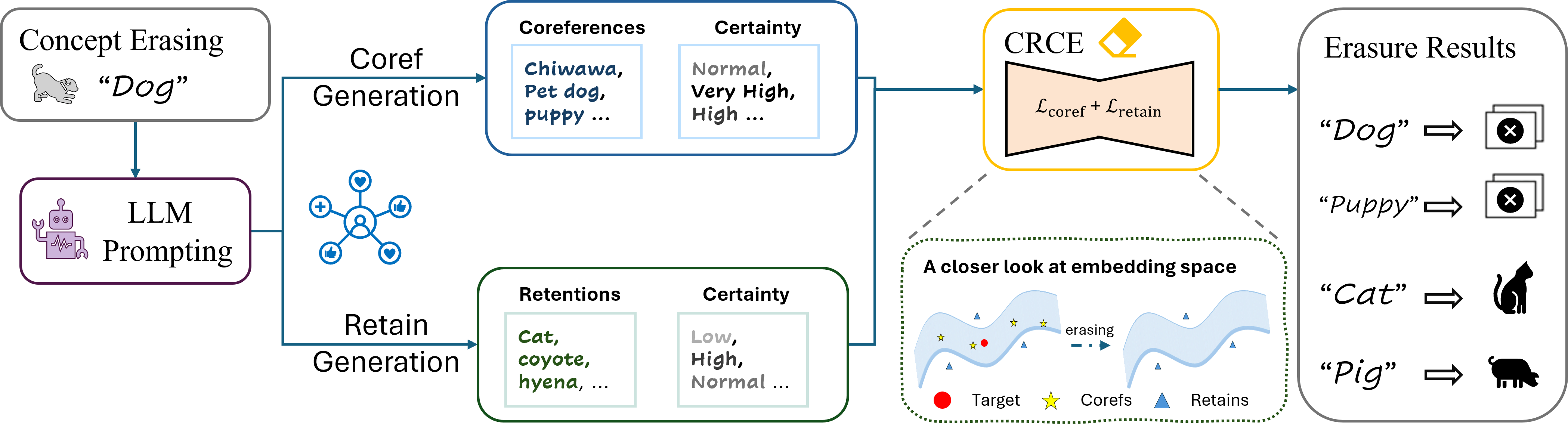}
   \caption{
   Overview of our proposed CRCE method.
   Our method erases a target concept (\eg ``dog'') while preserving unrelated concepts. 
   Using LLM prompting, we generate corefs (\eg ``Chihuahua'', ``puppy'') and retains (\eg ``cat'', ``coyote''), each assigned a certainty score. 
   The CRCE loss optimizes both coref erasure and retain preservation, adjusting the embedding space to minimize unintended removals. 
   The final erasure results show that the target (``dog'') and its coref terms (``puppy'') are erased, while unrelated concepts (``cat'', ``pig'') remain intact, ensuring effective and controlled concept erasing. 
   }
   \label{fig:4}
\end{figure*}

To tackle both under- and over-erasure, our driving hypothesis is that these issues stem from how concepts are structured within the CLIP
%~\cite{radford2021learning}
semantic space. At the core of our approach is the identification of coreferential concepts (corefs) and retain concepts (retains), which more accurately reflect semantic relationships. 
Rather than relying on naive spherical assumptions around target concepts, we leverage LLMs to guide the discovery of corefs that lie on the same semantic manifold as the target, and retain concepts that, while may not be nearby in embedding space, should be preserved. 
This manifold-aware perspective enables more precise and controlled concept erasure. Building upon the ESD framework, we introduce two key innovations: (1) an LLM-guided mechanism for discovering coref and retain concepts, and (2) a synthetic dataset, \emph{CorefConcept}, to support this process. Additionally, we propose a loss function that jointly removes target concepts and their corefs while preserving unrelated retains. Together, these contributions form our Coref-Retain Concept Erasing (CRCE) approach, a robust and comprehensive solution to mitigate both under-erasure and over-erasure. 
Fig.~\ref{fig:4} illustrates the complete pipeline of our proposed CRCE method. The process begins with a target concept for erasure, which is passed to an LLM to generate appropriate corefs and retains with associated certainty levels. These generated concepts are then integrated into our loss function, which guides the model to effectively remove the target concept and its corefs while preserving retains.

\subsection{LLM-based Corefs and Retains Generation}
\label{sec:corefconcept}

We first formalise our definition of coreferences, as prior works may use the term with varying meanings. 
We define coreferences (corefs) as a type of concept that encompasses the original concept itself, including its \textbf{synonyms and all possible expressions} that may lead to the same conceptual understanding. Although some concepts might lack direct synonyms, they can have coreferential expressions. For instance, corefs for ``Mickey Mouse'' could be ``The first mouse character by Walt Disney''. ``The 45th President of the United States'' could be an appropriate coref for ``Donald Trump'';
For retained concepts, we take a more nuanced approach, beyond simply selecting arbitrary unrelated concepts. We define retains as concepts that share semantic proximity to the target concept without being coreferential. These concepts typically exist in the same categorical neighbourhood, but are fundamentally distinct entities. 
Importantly, we select retains that are close enough in the embedding space to potentially be affected by over-erasure, making them valuable indicators of erasure precision. For example, when erasing ``dog'', concepts such as ``cat'' and ``pig'' serve as ideal retains because they share categorical similarities (mammals) while being distinct in both taxonomy and human perception. This careful selection of semantically adjacent but distinct concepts allows us to effectively mitigate over-erasure.
Note that using completely unrelated concepts (such as ``airplane'' when erasing ``dog''    ) would be trivially easy to retain and would fail to meaningfully evaluate the model's ability to avoid over-erasure, as such semantically distant concepts are unlikely to be affected even by imprecise erasure methods.

As illustrated in Fig.~\ref{fig:3}, randomly sampled embeddings often fail to accurately identify the correct corefs and retains, which are crucial for effective concept erasure. 
However, manually determining these corefs and retains for each concept is both time-consuming and labour-intensive.
To address this challenge and systematically identify appropriate coreferential and retain concepts, we harness the comprehensive semantic knowledge embedded in LLMs. These models excel at understanding nuanced relationships between concepts, making them ideal for generating semantically accurate concept associations that would be impractical to define manually at scale.

Our approach employs a structured prompting methodology that guides the LLM to generate distinct sets of concept relationships. For each target concept,
%(categorised as object, intellectual property, or celebrity), 
the LLM produces a ranked list of coreferential concepts that maintain visual and semantic connections to the target. These coreferentials range from direct synonyms to more loosely associated terms, each assigned a certainty level (``Very High'', ``High'', ``Normal'', ``Low'' or ``Very Low'') that quantifies the strength of their relationship to the target concept. 

The prompt encourages corefs to prioritise synonyms and high-precision semantic equivalents, while retains focus on concepts that share categorical proximity without being identical. Our carefully crafted prompts explicitly discourage vague descriptions or unrelated concepts, promoting specificity and relevance in the generated associations. In cases where the LLM generates inappropriate concept relationships, the system supports iterative refinement through multi-round conversations, allowing domain experts to adjust and optimise the generated corefs and retains as needed.
For a more detailed prompt template see Appendix~\ref{sec:app_prompt}.
% \yuyang{See Appendix~\ref{sec:app_prompt} for more details.}

\subsection{CorefConcept: A Dataset}

We employ the data-processing method described above to create a dataset \textit{CorefConcept} with  three categories: objects, Intellectual Property (IP), and celebrities.
We included CIFAR-10 class labels and some concepts that may be ambiguous (\eg ``bat (animal)'' vs ``bat (sports equipment)'') to form the object category. IP and celebrity categories were generated using the ChatGPT-o1 reasoning assistant~\cite{jaech2024openai}. Specifically, for celebrities, we provided the prompt: ``\textit{Provide a list of well-known celebrities, including various gender, ethnicity, age, etc.}'' For IP items, we prompted: ``\textit{Provide a list of well-known IP characters from different media, such as movies, books, cartoon, games, etc.}''
The dataset contains 20 object concepts, 15 IP characters and 15 celebrities.
Each concept was augmented with a list of 15 corefs and a list of 15 retains. We randomly selected 10 corefs per concept for the training set, with the remaining 5 used for the test set. 
Full details, including prompts used and dataset samples, are available in Appendices~\ref{sec:app_prompt_task} and~\ref{sec:dataset}, respectively.
This dataset provides a valuable resource for studying coreference and disambiguation in vision-language models, supporting research on reasoning and representation learning across diverse and ambiguous concepts.
\footnote{We will release the dataset and source code upon acceptance.}

\subsection{Coreference and Retention-Aware Loss} %Function}
\label{sec:loss}

We integrate generated coref and retain samples into the loss function to improve the concept erasing performance.
We add terms to the ESD loss to explicitly account for coref and retain concepts with their associated certainty levels. The loss function is defined as:
\begin{equation*}
    \begin{aligned}
      \mathcal{L}=&\mathcal{L}_{\text{ESD}} +
      \frac{1}{M}\sum_{c'\in \text{Coref}} \mathcal{C}_{c'} \cdot \left[\|\epsilon_\theta(c')-\epsilon_{\theta^*}^{\text{anchor}}(c)\|^2_2\right] 
      +\frac{1}{N}\sum_{r\in \text{Retain}} \mathcal{C}_{r} \cdot \left[\|\epsilon_\theta(r)-\epsilon_{\theta^*}(r)\|^2_2\right],
    \end{aligned}
%  \label{eq:Ours}
\end{equation*}
where $\mathcal{L}_{\text{ESD}}$ represents the ESD loss from Eq.~\ref{eq:ESD_diffusion}.
We introduce two additional terms that incorporate coref concepts $c'$ and retain concepts $r$ with each weighted by their certainty levels $\mathcal{C}_{c'}$ and $\mathcal{C}_{r}$ (with the certainty of each concept determined by the LLM). 
For each iteration, we randomly sample $M$ corefs and $N$ retains from the training corefs and the training retains.
The discrete certainty levels are $\{1.0,0.8,0.6,0.4,0.2\}$
corresponding from ``Very High'' to ``Very Low''.
Hence, concepts with higher certainty contribute more significantly to the loss, thereby prioritising alignment with target corefs and retains. 

\section{Experiments}
\label{sec:exp}

\subsection{Experimental Setup}
\label{sec:setup}

\textbf{Approaches for comparison}: We treat ESD-x~\cite{gandikota2023erasing} as a baseline method in the fine-tuning concept erasure paradigm. We also compare our method to other state-of-the-art methods: FMN~\cite{zhang2024forget}, UCE~\cite{gandikota2024unified}, SPM~\cite{lyu2024one}, 
%EUC~\cite{bui2024erasing}, 
Receler~\cite{huang2024receler}, MACE~\cite{lu2024mace}, and RealEra~\cite{liu2024realera}.
%\footnote{Original code unavailable - method was reimplemented locally.}
See Appendix~\ref{sec:implementation} for more information about implementation details.

\noindent\textbf{Evaluation metrics}: 
Conventional concept erasure evaluations use classifiers (\eg ResNet~\cite{he2016deep}) or CLIP Score~\cite{hessel2021clipscore} to assess erasure and preservation accuracy. However, classifiers face out-of-distribution (OOD) issues, and CLIP similarities may not reliably reflect semantic alignment with concepts~\cite{li2024erroneous} (see Sec.~\ref{sec:clip}). To address this, we propose an alternative evaluation protocol leveraging VLMs, which are well-suited for recognizing complex visual patterns across diverse domains.
%Conventional methods for evaluating concept erasure performance include the use of classifiers (\eg ResNet~\cite{he2016deep}) or CLIP Score~\cite{hessel2021clipscore} to determine the erasure and preservation accuracies of a post-erasure model. However, classifier-based methods suffer from out-of-distribution (OOD) issues, while CLIP similarities may fail to capture the direct semantic relationship between images and concepts directly~\cite{li2024erroneous} as we mentioned in Sec.~\ref{sec:clip}.
%In contrast, VLMs typically benefit from broad exposure to information across diverse domains and demonstrate high efficacy in recognising complex patterns in images. This motivates us to propose an alternative VLM-based evaluation protocol for the task.
Specifically, we select the open-source Qwen2 VL~\cite{wang2024qwen2} to instantiate performance evaluation. We prompt the model using the prompt: ``\textit{Observe and describe the image and check whether it has the same concept as} \texttt{<X>}'',
where the variable is instantiated by an instance of an object, character, or celebrity identity.
%\yuyang{.\st{; to test instances concerning objects, intellectual property, and celebrities, respectively.}}
We restrict the model to provide binary output (\textit{yes} or \textit{no}), as a % functioning as a 
straightforward classifier. 

We assess concept erasure performance with metrics that evaluate both erasure effectiveness and concept preservation: 
$Acc_{\text{U}}\downarrow$ measures how effectively the primary target concept is removed; 
$Acc_{\text{C}}^{\text{train}}\downarrow$ captures the erasure of coref concepts seen during fine-tuning, which are semantically related and should also be erased; 
To assess generalisation, $Acc_{\text{C}}^{\text{test}}\downarrow$ evaluates erasure accuracy on unseen coref concepts; 
On the preservation side, $Acc_{\text{R}}^{\text{train}}\uparrow$ reflects how well unrelated but semantically close concepts are retained during training, 
while $Acc_{\text{R}}^{\text{test}}\uparrow$ measures the model's ability to generalise this preservation to unseen retain concepts.
% \begin{itemize}
%     \item $Acc_{\text{U}}\downarrow$: Erasure accuracy on the primary target concept.
%     \item $Acc_{\text{C}}^{\text{train}}\downarrow$: Erasure accuracy on coref concepts used during fine-tuning. These semantically related concepts should also be erased.
%     \item $Acc_{\text{C}}^{\text{test}}\downarrow$: Erasure accuracy on held-out coref concepts not seen during training, measuring generalisation.
%     \item $Acc_{\text{R}}^{\text{train}}\uparrow$: Preservation accuracy on retain concepts used during fine-tuning. These concepts should remain intact despite semantic proximity.
%     \item $Acc_{\text{R}}^{\text{test}}\uparrow$: Preservation accuracy on held-out retain concepts, evaluating preservation generalisation.
% \end{itemize}
This train/test split approach assesses both performance on concepts used during optimisation and generalisation to unseen concepts. An effective method will have low values for erasure metrics and high values for preservation metrics, suggesting better discrimination between related and unrelated concepts.

\begin{table*}[t]
  \centering
  \caption{Quantitative comparison of concept erasure methods across Object, Intellectual Property (IP), and Celebrity categories. The top 3 accuracy is marked in \textbf{bold}, \underline{underlined}, and \textit{italics}, respectively. Our method (CRCE) consistently achieves the most balanced accuracy, effectively removing the target concepts and their coref ($Acc_{\text{U}}$, $Acc_{C}^{\text{test}}$) while preserving unrelated retained concepts ($Acc_{R}^{\text{test}}$). Empirically, we set the number of corefs and retains to $5$ and $3$, according to Table~\ref{tab:merged_accuracy} in Appendix \ref{sec:app_abla:num}.}
  \resizebox{\textwidth}{!}{
    \begin{tabular}{l|rrrrr|rrrrr|rrrrr}
    %\toprule
    \hline\hline
           & \multicolumn{5}{c|}{Object}           & \multicolumn{5}{c|}{Intellectual Property (IP)}               & \multicolumn{5}{c}{Celebrity} \\
\cline{2-16}
& \multicolumn{1}{l}{$Acc_{\text{U}}\downarrow$} & \multicolumn{1}{l}{$Acc_{\text{C}}^{\text{train}}\downarrow$} & \multicolumn{1}{l}{$Acc_{\text{C}}^{\text{test}}\downarrow$} & \multicolumn{1}{l}{$Acc_{\text{R}}^{\text{train}}\uparrow$} & \multicolumn{1}{l|}{$Acc_{\text{R}}^{\text{test}}\uparrow$} & \multicolumn{1}{l}{$Acc_{\text{U}}\downarrow$} & \multicolumn{1}{l}{$Acc_{\text{C}}^{\text{train}}\downarrow$} & \multicolumn{1}{l}{$Acc_{\text{C}}^{\text{test}}\downarrow$} & \multicolumn{1}{l}{$Acc_{\text{R}}^{\text{train}}\uparrow$} & \multicolumn{1}{l|}{$Acc_{\text{R}}^{\text{test}}\uparrow$} & \multicolumn{1}{l}{$Acc_{\text{U}}\downarrow$} & \multicolumn{1}{l}{$Acc_{\text{C}}^{\text{train}}\downarrow$} & \multicolumn{1}{l}{$Acc_{\text{C}}^{\text{test}}\downarrow$} & \multicolumn{1}{l}{$Acc_{\text{R}}^{\text{train}}\uparrow$} & \multicolumn{1}{l}{$Acc_{\text{R}}^{\text{test}}\uparrow$} \\
    %\midrule
    %\midrule
    \hline
    SD~\cite{rombach2022high}& 83.33 & 87.22 & 91.60 & 81.17 & 80.63 & 88.33 & 37.46 & 30.13 & 86.80 & 86.75 & 90.90 & 17.12 & 17.94 & 92.64 & 93.24  \\
    %\cmidrule{1-16} 
    \hline
        %& ESD-U~\cite{gandikota2023erasing} & \textbf{1.17} & \textbf{6.04} & \textbf{6.94} & 23.93 & 28.79 & \textbf{0.00} & \textbf{1.47} & \textbf{2.67} & 24.67 & 28.31 & \textbf{0.00} & \underline{1.73} & \textbf{1.33} & 7.33 & 6.67 \\
        ESD-x~\cite{gandikota2023erasing} & 3.33 & 37.72 & 39.92 & 73.82 & 69.89 & 3.33 & 10.67 & 7.73 & 62.80 & 71.16 & \textit{1.81} & 11.09 & 10.76 & 72.9 & 68.37 \\
        FMN~\cite{zhang2024forget} & 4.00 &	50.78	&50.45	&30.06	&25.92&	4.93	&4.45	&\textit{3.89}	&2.88	&2.51	&9.2	&4.38	&4.00	&7.92	&5.23\\
        UCE~\cite{gandikota2024unified} & \underline{1.27} & \underline{8.52}& \underline{8.40} & 17.91 & 14.58& \textit{1.17} & \underline{1.38} &6.57 & 1.01 & 1.02 &\underline{1.47} & 3.14& 3.03& 5.83& 5.00\\
        SPM~\cite{lyu2024one} & 36.00 & 31.04 & 33.29 & 14.57 & 15.45 & 16.86 & 9.52 & 10.13 & 5.87 & 8.05 & 16.85 & 15.78 & 19.41 & 33.94 & 34.35\\	
        % EUC~\cite{bui2024erasing}\\
        Receler~\cite{huang2024receler} &2.40	& \textbf{6.53}	& \textbf{7.09}	&11.68	&8.71	&\underline{0.27}	&\textbf{0.77}	&\textbf{1.74}	&0.87	&0.91	&0.53	&\textit{2.87}	&1.76	&4.11	&4.96\\
        MACE~\cite{lu2024mace} & \textit{2.10} & \textit{12.10} & \textit{9.05} & 62.73 & 68.84  & 1.33 & \textit{3.33} & \underline{3.73} & 44.53 & 3.73 & \textbf{0.00} & 3.33 & \underline{2.40} & 29.73 & 30.40   \\
        RealEra~\cite{liu2024realera} & 48.88 & 70.49 & 72.60 & \textbf{88.92} & \textit{77.15} & 3.33 & 16.67 & 14.66 & \underline{77.73} & \textbf{81.81} & 12.72 & 11.78 & 11.53 & \underline{82.19} & \underline{85.40}  \\
    %\cmidrule{1-16}          
    \hline
        CRCE-Sphere & 26.67 & 58.11 & 62.40 & \underline{86.57} & \textbf{84.84} & 20.00 & 13.73 & 12.00 & \textit{75.20} & \textit{79.47} & 7.27 & 11.78 & 11.53 & \underline{82.19} & \textit{82.70} \\
        %CRCE-Plain & \\
        CRCE-Fixed & 8.75 & 19.27 & 32.21 & 78.23 & 72.39 & 7.66 & 4.80 & 4.26 & 67.94 & 69.47 & 5.57 & \textbf{1.36} & \textit{2.59} & \textit{81.54} & 79.45 \\
        CRCE & \textbf{1.25} & 12.81 & \underline{26.31} & \textit{85.20} & \underline{79.56} & \textbf{0.00} & 6.27 & 7.46 & \textbf{78.80} & \underline{80.25} & \textit{1.81} & \underline{2.04} & \textbf{1.29} & \textbf{87.35} & \textbf{87.02} \\
    \bottomrule
    \end{tabular}%
    }
  \label{tab:exp}%
\end{table*}%

\subsection{Quantitative Results}
\label{sec:results}

\begin{figure*}[t]
  \centering
  \includegraphics[width=\linewidth]{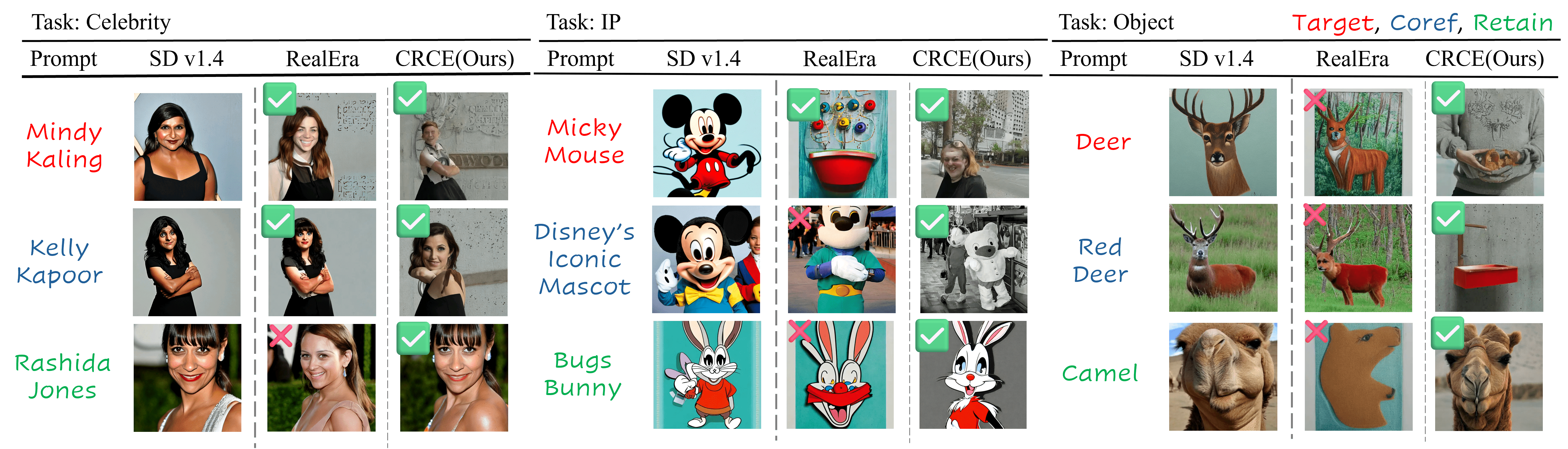}
   \caption{
   Comparison of concept erasure effectiveness between RealEra~\cite{liu2024realera} and our method. ``Mindy Kaling'' (celebrity), ``Micky Mouse'' (IP), and ``Deer'' (object) are targeted for removal along with their corefs. CRCE successfully erases corefs while accurately retaining related yet distinct entities, demonstrating superior precision compared to RealEra. 
   %Appendix~\ref{sec:more}
   %provides comprehensive visuals across all comparison methods. %presents more comprehensive visuals showing more samples and methods.
   }
   \label{fig:5}
\end{figure*}

Table~\ref{tab:exp} shows a clear pattern: existing erasure methods excel on one side of the coref-retain trade-off but stumble on the other, whereas \textbf{CRCE balances both}. 
From $Acc_{\text{U}}$, $Acc_{\text{C}}^{\text{test}}$ and $Acc_{\text{R}}^{\text{test}}$ for Object, we can observe that RealEra, which retain from a random CLIP-sphere of neighbours, preserves unrelated concepts best, \eg it keeps retain accuracy at 77.15\%, yet leaves the target itself recognisable in 48.88\% of cases and coreferential variants in 72.60\%, underscoring severe under-erasure. Receler achieves the opposite extreme: by steering cross-attention to a single anchor it drives target accuracy to 2.40\% and corefs to 7.09\%, but the collateral damage is strong, with retain accuracy collapsing to 8.71\% on the test split, showing serious over-erasure. 
Adapter-based SPM (latent anchoring) and attention-reweighting FMN attempt to strike middle ground—where retain accuracy rises to 15.45\% to 25.92\%, respectively, yet more than one-third of corefs are not erased, while robustness-oriented UCE wipes out almost everything, leaving retain accuracy below 15\%.
MACE does a relatively good job on both remove target and corefs, while preserving two thirds of retains.
In contrast, CRCE drives the accuracy of under-erasure to single digits and keeps unrelated prompts above 79\% in every domain: on Objects it lowers $Acc_{\text{U}}$ to 1.25\% and coref accuracy to 12.8\% while still scoring 79.6\% on retains; on Intellectual-Property (IP) the corresponding numbers are 0\%, 7.46\%, and 80.25\%; on Celebrities 1.81\%, 1.29\%, and 87.02\%. 
These quantitative margins match visual inspection (see Appendix~\ref{sec:more} for more details): rivals that protect image quality often leave clear visual traces of the target (\eg object ``horse'' cannot be erased using FMN or RealEra), whereas those that purge aggressively distort or erase desirable details (\eg SPM and MACE's scenes turn to texture, Receler's prompts yield pure colors).
CRCE alone removes every instance together with their corefs while leaving retains untouched, empirically demonstrating that the LLM-derived coref/retain manifolds and the certainty-weighted objective deliver the sought-after balance between erasure precision and content preservation.
Fig.~\ref{fig:5} presents the selected visual comparisons between our method and RealEra across object, celebrity, and IP categories. Comprehensive results comparing all methods are presented at Appendix~\ref{sec:more}.

%Fig.~\ref{fig:5} shows visual examples of object, celebrity, and IP categories. Appendix~\ref{sec:more} presents more comprehensive visual results.

\subsection{Ablation Study: Variations of CRCE}
\label{sec:abla:var}
To better understand the impact of different sampling strategies on concept erasure performance, We ablated our sampling strategy with two variants:
\textbf{CRCE-Sphere} employs RealEra’s CLIP-sphere sampling strategy but uses our loss. Results show that it reduces $Acc_{\text{U}}$ on Objects from 48.88\% to 26.67\% and lifts retain accuracy from 72.60\% to 84.84\%, confirming the effectiveness of our loss even under Euclidean assumptions.
\textbf{CRCE-Fixed} trains on one static coref/retain list, giving consistent but narrow supervision. 
\textbf{CRCE} (our full method) randomly samples new coref/retain prompts at each step, covering the manifold more completely and therefore beats CRCE-Sphere/Fixed across all metrics as shown in Table~\ref{tab:exp}.
The gaps between \textbf{Sphere} $\rightarrow$ \textbf{CRCE-Fixed} $\rightarrow$ \textbf{CRCE} underscore that (i) CLIP space is not isotropic and (ii) dynamic sampling prevents over-fitting to a local concept region.

The results of other two ablation studies on \textit{the coref and retain certainty} and \textit{their number needed} are available in Appendix~\ref{sec:app_abla}. These ablations show that structured certainty scores, especially for corefs, are crucial for effective and robust concept erasure, as random or uniform certainty leads to significant performance degradation. Additionally, using a moderate number of corefs and retains (\eg $M=5$, $N=3$) achieves the best balance across metrics, while overly large values introduce semantic interference and reduce generalization performance.

\section{Conclusions}

We introduce CRCE, a novel concept erasure approach for T2I diffusion models that tackles under- and over-erasure by leveraging LLMs to identify semantically related (coref) and unrelated (retain) concepts, enabling more precise erasure compared to distance-based sampling methods. 
Our \textit{CorefConcept} dataset and tailored loss enable fine-grained erasure and preservation. 
%CRCE outperforms prior methods across diverse tasks and is adaptable beyond the ESD backbone. Not restricted to the ESD backbone and can integrate with any diffusion-based concept erasure method.
CRCE outperforms prior methods across diverse tasks and is not restricted to the ESD backbone but can seamlessly integrate with any diffusion-based concept erasure framework.
Limitations such as difficult erasure in concept entanglement are addressed in Appendix~\ref{sec:limitations}.
Future work may extend CRCE to stylistic or explicit concepts, potentially incorporating human feedback to refine semantic relationships.

%We introduce CRCE, a novel concept erasure approach for T2I diffusion models that addresses under-erasure and over-erasure challenges. Our key innovation uses LLMs to strategically identify points on the concept manifold (corefs) and off the manifold (retains), enabling more precise erasure compared to Euclidean-based sampling methods. The \textit{CorefConcept} dataset maps semantic relationships between concepts with certainty metrics, while our loss function effectively balances erasure and preservation. Our approach outperforms existing methods across object, celebrity, and IP erasure tasks. Importantly, CRCE isn't restricted to the ESD backbone and can integrate with any text-embedding-based concept erasure method. Future work could extend CRCE to more concepts such as artistmore concepts such as artistic styles or explicit content, potentially incorporating human-in-the-loop mechanisms to refine concept relationships.
%\yuyang{more concepts such as artistmore concepts such as artistic styles or explicit content, \st{other models and domain-specific applications,}} 

\bibliography{egbib}

\begin{thebibliography}{43}
\providecommand{\natexlab}[1]{#1}
\providecommand{\url}[1]{\texttt{#1}}
\expandafter\ifx\csname urlstyle\endcsname\relax
  \providecommand{\doi}[1]{doi: #1}\else
  \providecommand{\doi}{doi: \begingroup \urlstyle{rm}\Url}\fi

\bibitem[Balaji et~al.(2022)Balaji, Nah, Huang, Vahdat, Song, Zhang, Kreis, Aittala, Aila, Laine, et~al.]{balaji2022ediff}
Yogesh Balaji, Seungjun Nah, Xun Huang, Arash Vahdat, Jiaming Song, Qinsheng Zhang, Karsten Kreis, Miika Aittala, Timo Aila, Samuli Laine, et~al.
\newblock ediff-i: Text-to-image diffusion models with an ensemble of expert denoisers.
\newblock \emph{arXiv preprint arXiv:2211.01324}, 2022.

\bibitem[Bui et~al.(2025)Bui, Vu, Vuong, Le, Montague, Abraham, Kim, and Phung]{bui2025fantastic}
Anh Bui, Trang Vu, Long Vuong, Trung Le, Paul Montague, Tamas Abraham, Junae Kim, and Dinh Phung.
\newblock Fantastic targets for concept erasure in diffusion models and where to find them.
\newblock \emph{arXiv preprint arXiv:2501.18950}, 2025.

\bibitem[Chang et~al.(2023)Chang, Zhang, Barber, Maschinot, Lezama, Jiang, Yang, Murphy, Freeman, Rubinstein, et~al.]{chang2023muse}
Huiwen Chang, Han Zhang, Jarred Barber, AJ~Maschinot, Jose Lezama, Lu~Jiang, Ming-Hsuan Yang, Kevin Murphy, William~T Freeman, Michael Rubinstein, et~al.
\newblock Muse: Text-to-image generation via masked generative transformers.
\newblock In \emph{Proceedings of the 40th International Conference on Machine Learning (ICML)}, 2023.

\bibitem[Chowdhury et~al.(2025)Chowdhury, Dubey, Beirami, Kidambi, Monath, Ahmed, and Chaturvedi]{chowdhury2025fundamental}
Somnath Basu~Roy Chowdhury, Avinava Dubey, Ahmad Beirami, Rahul Kidambi, Nicholas Monath, Amr Ahmed, and Snigdha Chaturvedi.
\newblock Fundamental limits of perfect concept erasure.
\newblock \emph{arXiv preprint arXiv:2503.20098}, 2025.

\bibitem[Fan et~al.(2023)Fan, Liu, Zhang, Wong, Wei, and Liu]{fan2023salun}
Chongyu Fan, Jiancheng Liu, Yihua Zhang, Eric Wong, Dennis Wei, and Sijia Liu.
\newblock Salun: Empowering machine unlearning via gradient-based weight saliency in both image classification and generation.
\newblock In \emph{ICLR}, 2023.

\bibitem[Fu et~al.(2023)Fu, Tamir, Sundaram, Chai, Zhang, Dekel, and Isola]{fu2023dreamsim}
Stephanie Fu, Netanel Tamir, Shobhita Sundaram, Lucy Chai, Richard Zhang, Tali Dekel, and Phillip Isola.
\newblock Dreamsim: Learning new dimensions of human visual similarity using synthetic data.
\newblock \emph{NeurIPS}, 2023.

\bibitem[Gandikota et~al.(2023)Gandikota, Materzynska, Fiotto-Kaufman, and Bau]{gandikota2023erasing}
Rohit Gandikota, Joanna Materzynska, Jaden Fiotto-Kaufman, and David Bau.
\newblock Erasing concepts from diffusion models.
\newblock In \emph{ICCV}, pages 2426--2436, 2023.

\bibitem[Gandikota et~al.(2024)Gandikota, Orgad, Belinkov, Materzy{\'n}ska, and Bau]{gandikota2024unified}
Rohit Gandikota, Hadas Orgad, Yonatan Belinkov, Joanna Materzy{\'n}ska, and David Bau.
\newblock Unified concept editing in diffusion models.
\newblock In \emph{WACV}, pages 5111--5120, 2024.

\bibitem[Gong et~al.(2024)Gong, Chen, Wei, Chen, and Jiang]{gong2024reliable}
Chao Gong, Kai Chen, Zhipeng Wei, Jingjing Chen, and Yu-Gang Jiang.
\newblock Reliable and efficient concept erasure of text-to-image diffusion models.
\newblock In \emph{European Conference on Computer Vision}, pages 73--88. Springer, 2024.

\bibitem[He et~al.(2016)He, Zhang, Ren, and Sun]{he2016deep}
Kaiming He, Xiangyu Zhang, Shaoqing Ren, and Jian Sun.
\newblock Deep residual learning for image recognition.
\newblock In \emph{Proceedings of the IEEE conference on computer vision and pattern recognition}, pages 770--778, 2016.

\bibitem[Hessel et~al.(2021)Hessel, Holtzman, Forbes, Bras, and Choi]{hessel2021clipscore}
Jack Hessel, Ari Holtzman, Maxwell Forbes, Ronan~Le Bras, and Yejin Choi.
\newblock Clipscore: A reference-free evaluation metric for image captioning.
\newblock \emph{arXiv preprint arXiv:2104.08718}, 2021.

\bibitem[Hu et~al.(2022)Hu, Shen, Wallis, Allen-Zhu, Li, Wang, Wang, Chen, et~al.]{hu2022lora}
Edward~J Hu, Yelong Shen, Phillip Wallis, Zeyuan Allen-Zhu, Yuanzhi Li, Shean Wang, Lu~Wang, Weizhu Chen, et~al.
\newblock Lora: Low-rank adaptation of large language models.
\newblock \emph{ICLR}, 1\penalty0 (2):\penalty0 3, 2022.

\bibitem[Huang et~al.(2024)Huang, Chang, Tsai, Lai, Yang, and Wang]{huang2024receler}
Chi-Pin Huang, Kai-Po Chang, Chung-Ting Tsai, Yung-Hsuan Lai, Fu-En Yang, and Yu-Chiang~Frank Wang.
\newblock Receler: Reliable concept erasing of text-to-image diffusion models via lightweight erasers.
\newblock In \emph{European Conference on Computer Vision}, pages 360--376. Springer, 2024.

\bibitem[Hunter(2023)]{hunter2023ai}
Tatum Hunter.
\newblock Ai porn is easy to make now. for women, that's a nightmare.
\newblock \emph{The Washington Post}, 2023.

\bibitem[Jaech et~al.(2024)Jaech, Kalai, Lerer, Richardson, El-Kishky, Low, Helyar, Madry, Beutel, Carney, et~al.]{jaech2024openai}
Aaron Jaech, Adam Kalai, Adam Lerer, Adam Richardson, Ahmed El-Kishky, Aiden Low, Alec Helyar, Aleksander Madry, Alex Beutel, Alex Carney, et~al.
\newblock Openai o1 system card.
\newblock \emph{arXiv preprint arXiv:2412.16720}, 2024.

\bibitem[Jiang et~al.(2023)Jiang, Brown, Cheng, Khan, Gupta, Workman, Hanna, Flowers, and Gebru]{jiang2023ai}
Harry~H Jiang, Lauren Brown, Jessica Cheng, Mehtab Khan, Abhishek Gupta, Deja Workman, Alex Hanna, Johnathan Flowers, and Timnit Gebru.
\newblock Ai art and its impact on artists.
\newblock In \emph{AAAI}, pages 363--374, 2023.

\bibitem[Kim and Qi(2025)]{kim2025comprehensive}
Changhoon Kim and Yanjun Qi.
\newblock A comprehensive survey on concept erasure in text-to-image diffusion models.
\newblock \emph{arXiv preprint arXiv:2502.14896}, 2025.

\bibitem[Kumari et~al.(2023)Kumari, Zhang, Wang, Shechtman, Zhang, and Zhu]{kumari2023ablating}
Nupur Kumari, Bingliang Zhang, Sheng-Yu Wang, Eli Shechtman, Richard Zhang, and Jun-Yan Zhu.
\newblock Ablating concepts in text-to-image diffusion models.
\newblock In \emph{CVPR}, pages 22691--22702, 2023.

\bibitem[Lab(2024)]{blackforestlab_2024}
Black~Forest Lab.
\newblock Announcing flux1.1 [pro] and the bfl api, Oct 2024.
\newblock URL \url{https://blackforestlabs.ai/announcing-flux-1-1-pro-and-the-bfl-api/}.

\bibitem[Li et~al.(2024)Li, Koh, and Du]{li2024erroneous}
Siting Li, Pang~Wei Koh, and Simon~Shaolei Du.
\newblock On erroneous agreements of clip image embeddings.
\newblock \emph{arXiv preprint arXiv:2411.05195}, 2024.

\bibitem[Liu et~al.(2024{\natexlab{a}})Liu, Khakzar, Gu, Chen, Torr, and Pizzati]{liu2024latent}
Runtao Liu, Ashkan Khakzar, Jindong Gu, Qifeng Chen, Philip Torr, and Fabio Pizzati.
\newblock Latent guard: a safety framework for text-to-image generation.
\newblock In \emph{ECCV}, pages 93--109. Springer, 2024{\natexlab{a}}.

\bibitem[Liu et~al.(2024{\natexlab{b}})Liu, An, Zhang, Li, Wu, Gu, Lin, and Wang]{liu2024realera}
Yufan Liu, Jinyang An, Wanqian Zhang, Ming Li, Dayan Wu, Jingzi Gu, Zheng Lin, and Weiping Wang.
\newblock Realera: Semantic-level concept erasure via neighbor-concept mining.
\newblock \emph{arXiv preprint arXiv:2410.09140}, 2024{\natexlab{b}}.

\bibitem[Lu et~al.(2024)Lu, Wang, Li, Liu, and Kong]{lu2024mace}
Shilin Lu, Zilan Wang, Leyang Li, Yanzhu Liu, and Adams Wai-Kin Kong.
\newblock Mace: Mass concept erasure in diffusion models.
\newblock In \emph{CVPR}, pages 6430--6440, 2024.

\bibitem[Luccioni et~al.(2023)Luccioni, Akiki, Mitchell, and Jernite]{luccioni2023stable}
Alexandra~Sasha Luccioni, Christopher Akiki, Margaret Mitchell, and Yacine Jernite.
\newblock Stable bias: Analyzing societal representations in diffusion models.
\newblock \emph{arXiv preprint arXiv:2303.11408}, 2023.

\bibitem[Lyu et~al.(2024)Lyu, Yang, Hong, Chen, Jin, He, Xue, Han, and Ding]{lyu2024one}
Mengyao Lyu, Yuhong Yang, Haiwen Hong, Hui Chen, Xuan Jin, Yuan He, Hui Xue, Jungong Han, and Guiguang Ding.
\newblock One-dimensional adapter to rule them all: Concepts diffusion models and erasing applications.
\newblock In \emph{Proceedings of the IEEE/CVF Conference on Computer Vision and Pattern Recognition}, pages 7559--7568, 2024.

\bibitem[Oktay et~al.()Oktay, Schlemper, Le~Folgoc, Lee, Heinrich, Misawa, Mori, McDonagh, Hammerla, Kainz, et~al.]{oktayattention}
Ozan Oktay, Jo~Schlemper, Loic Le~Folgoc, Matthew Lee, Mattias Heinrich, Kazunari Misawa, Kensaku Mori, Steven McDonagh, Nils~Y Hammerla, Bernhard Kainz, et~al.
\newblock Attention u-net: Learning where to look for the pancreas.
\newblock In \emph{Medical Imaging with Deep Learning}.

\bibitem[Poppi et~al.(2023)Poppi, Poppi, Cocchi, Cornia, Baraldi, and Cucchiara]{poppi2023removing}
Samuele Poppi, Tobia Poppi, Federico Cocchi, Marcella Cornia, Lorenzo Baraldi, and Rita Cucchiara.
\newblock Removing nsfw concepts from vision-and-language models for text-to-image retrieval and generation.
\newblock \emph{CoRR}, 2023.

\bibitem[Qu et~al.(2023)Qu, Shen, He, Backes, Zannettou, and Zhang]{qu2023unsafe}
Yiting Qu, Xinyue Shen, Xinlei He, Michael Backes, Savvas Zannettou, and Yang Zhang.
\newblock Unsafe diffusion: On the generation of unsafe images and hateful memes from text-to-image models.
\newblock In \emph{Proceedings of the 2023 ACM SIGSAC conference on computer and communications security}, pages 3403--3417, 2023.

\bibitem[Radford et~al.(2021)Radford, Kim, Hallacy, Ramesh, Goh, Agarwal, Sastry, Askell, Mishkin, Clark, et~al.]{radford2021learning}
Alec Radford, Jong~Wook Kim, Chris Hallacy, Aditya Ramesh, Gabriel Goh, Sandhini Agarwal, Girish Sastry, Amanda Askell, Pamela Mishkin, Jack Clark, et~al.
\newblock Learning transferable visual models from natural language supervision.
\newblock In \emph{Proceedings of the 38th International Conference on Machine Learning (ICML)}, pages 8748--8763. PMLR, 2021.

\bibitem[Ramesh et~al.(2022)Ramesh, Dhariwal, Nichol, Chu, and Chen]{ramesh2022hierarchical}
Aditya Ramesh, Prafulla Dhariwal, Alex Nichol, Casey Chu, and Mark Chen.
\newblock Hierarchical text-conditional image generation with clip latents.
\newblock \emph{arXiv preprint arXiv:2204.06125}, 1\penalty0 (2):\penalty0 3, 2022.

\bibitem[Rombach et~al.(2022)Rombach, Blattmann, Lorenz, Esser, and Ommer]{rombach2022high}
Robin Rombach, Andreas Blattmann, Dominik Lorenz, Patrick Esser, and Bj{\"o}rn Ommer.
\newblock High-resolution image synthesis with latent diffusion models.
\newblock In \emph{CVPR}, pages 10684--10695, 2022.

\bibitem[Roose(2022)]{roose2022ai}
Kevin Roose.
\newblock An ai-generated picture won an art prize. artists aren't happy.
\newblock \emph{The New York Times}, 2022.

\bibitem[Saharia et~al.(2022)Saharia, Chan, Saxena, Li, Whang, Denton, Ghasemipour, Gontijo~Lopes, Karagol~Ayan, Salimans, et~al.]{saharia2022photorealistic}
Chitwan Saharia, William Chan, Saurabh Saxena, Lala Li, Jay Whang, Emily~L Denton, Kamyar Ghasemipour, Raphael Gontijo~Lopes, Burcu Karagol~Ayan, Tim Salimans, et~al.
\newblock Photorealistic text-to-image diffusion models with deep language understanding.
\newblock \emph{NeurIPS}, 35:\penalty0 36479--36494, 2022.

\bibitem[Schramowski et~al.(2023)Schramowski, Brack, Deiseroth, and Kersting]{schramowski2023safe}
Patrick Schramowski, Manuel Brack, Bj{\"o}rn Deiseroth, and Kristian Kersting.
\newblock Safe latent diffusion: Mitigating inappropriate degeneration in diffusion models.
\newblock In \emph{CVPR}, pages 22522--22531, 2023.

\bibitem[Schuhmann et~al.(2022)Schuhmann, Beaumont, Vencu, Gordon, Wightman, Cherti, Coombes, Katta, Mullis, Wortsman, et~al.]{schuhmann2022laion}
Christoph Schuhmann, Romain Beaumont, Richard Vencu, Cade Gordon, Ross Wightman, Mehdi Cherti, Theo Coombes, Aarush Katta, Clayton Mullis, Mitchell Wortsman, et~al.
\newblock Laion-5b: An open large-scale dataset for training next generation image-text models.
\newblock \emph{NeurIPS}, 35:\penalty0 25278--25294, 2022.

\bibitem[Setty(2023)]{Setty2023ai}
Riddhi Setty.
\newblock Ai art generators hit with copyright suit over artists' images.
\newblock \emph{Bloomberg Law}, 2023.

\bibitem[Somepalli et~al.(2023)Somepalli, Singla, Goldblum, Geiping, and Goldstein]{somepalli2023diffusion}
Gowthami Somepalli, Vasu Singla, Micah Goldblum, Jonas Geiping, and Tom Goldstein.
\newblock Diffusion art or digital forgery? investigating data replication in diffusion models.
\newblock In \emph{CVPR}, pages 6048--6058, 2023.

\bibitem[Struppek et~al.(2023)Struppek, Hintersdorf, Friedrich, Schramowski, Kersting, et~al.]{struppek2023exploiting}
Lukas Struppek, Dom Hintersdorf, Felix Friedrich, Patrick Schramowski, Kristian Kersting, et~al.
\newblock Exploiting cultural biases via homoglyphs in text-to-image synthesis.
\newblock \emph{Journal of Artificial Intelligence Research}, 78:\penalty0 1017--1068, 2023.

\bibitem[Tsai et~al.(2023)Tsai, Hsu, Xie, Lin, Chen, Li, Chen, Yu, and Huang]{tsai2023ring}
Yu-Lin Tsai, Chia-Yi Hsu, Chulin Xie, Chih-Hsun Lin, Jia-You Chen, Bo~Li, Pin-Yu Chen, Chia-Mu Yu, and Chun-Ying Huang.
\newblock Ring-a-bell! how reliable are concept removal methods for diffusion models?
\newblock \emph{arXiv preprint arXiv:2310.10012}, 2023.

\bibitem[Wang et~al.(2024)Wang, Bai, Tan, Wang, Fan, Bai, Chen, Liu, Wang, Ge, et~al.]{wang2024qwen2}
Peng Wang, Shuai Bai, Sinan Tan, Shijie Wang, Zhihao Fan, Jinze Bai, Keqin Chen, Xuejing Liu, Jialin Wang, Wenbin Ge, et~al.
\newblock Qwen2-vl: Enhancing vision-language model's perception of the world at any resolution.
\newblock \emph{arXiv preprint arXiv:2409.12191}, 2024.

\bibitem[Zhang et~al.(2024{\natexlab{a}})Zhang, Wang, Xu, Wang, and Shi]{zhang2024forget}
Gong Zhang, Kai Wang, Xingqian Xu, Zhangyang Wang, and Humphrey Shi.
\newblock Forget-me-not: Learning to forget in text-to-image diffusion models.
\newblock In \emph{CVPRW}, pages 1755--1764, 2024{\natexlab{a}}.

\bibitem[Zhang et~al.(2024{\natexlab{b}})Zhang, Chen, Jia, Zhang, Fan, Liu, Hong, Ding, and Liu]{zhang2024defensive}
Yimeng Zhang, Xin Chen, Jinghan Jia, Yihua Zhang, Chongyu Fan, Jiancheng Liu, Mingyi Hong, Ke~Ding, and Sijia Liu.
\newblock Defensive unlearning with adversarial training for robust concept erasure in diffusion models.
\newblock In \emph{The Thirty-eighth Annual Conference on Neural Information Processing Systems}, 2024{\natexlab{b}}.

\bibitem[Zhang et~al.(2024{\natexlab{c}})Zhang, Jia, Chen, Chen, Zhang, Liu, Ding, and Liu]{zhang2024generate}
Yimeng Zhang, Jinghan Jia, Xin Chen, Aochuan Chen, Yihua Zhang, Jiancheng Liu, Ke~Ding, and Sijia Liu.
\newblock To generate or not? safety-driven unlearned diffusion models are still easy to generate unsafe images... for now.
\newblock In \emph{ECCV}, pages 385--403. Springer, 2024{\natexlab{c}}.

\end{thebibliography}

\clearpage
\newpage

\appendix
\addcontentsline{toc}{section}{Appendices}
% \onecolumn

\section{Additional Related Work}
\label{sec:app_related_work}

\subsection{Text-to-Image (T2I) Diffusion Models}
\label{sec:app_t2i}

Text-to-Image (T2I) generation is a branch of generative modeling that focuses on producing images from textual descriptions, where diffusion-based solutions currently dominate the field.
%Recent breakthroughs in model architectures and large-scale training have significantly enhanced the fidelity and coherence of generated images.
Stable Diffusion (SD)~\cite{rombach2022high} constitutes a highly %one of the most 
popular open-source latent diffusion model (LDM) that reduces computational costs by operating within a compressed latent space.
%Meanwhile, DALL·E 2~\cite{ramesh2022hierarchical} introduced a two-stage ``unCLIP'' pipeline, thereby improving sample diversity and artistic expressivity.  
eDiff-I~\cite{balaji2022ediff} uses an ensemble of specialised diffusion experts to handle highly compositional prompts, while Muse~\cite{chang2023muse} explores masked generative transformers for parallelised image synthesis.
Similarly, Flux~\cite{blackforestlab_2024} proposes an architecture that combines diffusion and autoregressive components to balance the generation speed with high-quality outputs. 
Emerging research on concept erasure addresses pressing concerns around privacy and content moderation, highlighting the ethical and legal dimensions of large-scale T2I models. %Overall, these developments emphasize the field's ongoing efforts to refine controllability, efficiency, and visual fidelity in T2I synthesis. 
We will take a more detailed review of LDM.
In our study, we build upon SD v1.4~\cite{rombach2022high} as the foundational model for implementing concept erasure algorithms, given its recent popularity and quasi benchmark status. 

%\subsubsection{Latent Diffusion Model (LDM)}
%\label{sec:app_LDM}
% Introduction of LD
LDM~\cite{rombach2022high}, typically have three main components: 1) a pre-trained vision autoencoder to compress high-dimensional image data into a low-dimensional latent representation,
% The auto encoder
The encoding network $\mathcal{E}(\cdot)$ maps an image $x$ to a latent variable $z$, and the decoding network $\mathcal{D}(\cdot)$ reconstructs the image from the latent space such that $\mathcal{D}(z)=\hat{x}\approx x$.
2) 
% The text condition
The text encoder transforms text prompts into conditioning vectors, allowing a condition over the generation process. The textual prompt $p$ is embedded as $c=\mathcal{T}(p)$, where $\mathcal{T}$ is a text encoder, typically CLIP~\cite{radford2021learning}.
% The LD model
3) A latent diffusion model for iterative denoising that accept text embedding $c$ through the cross-attention layers in the core generative process, governed by a U-Net-based~\cite{oktayattention} LDM that progressively refines noisy latent representations towards high-fidelity outputs along the diffusion trajectory.

\subsection{Concept Erasure}
\label{sec:app_ce}

The primary objective of concept erasure is to condition the model such that it inherently removes the concept in response to undesired prompts. This can be accomplished through the following paradigms.
%Concept erasure can be achieved through various paradigms.

\textbf{Closed-form Model Editing} offers a direct mathematical update to model parameters without requiring iterative training.
A general framework for closed-form model editing utilises least squares-based optimisation, primarily focusing on the key and value projection matrices within the LDM cross-attention module. 
%TIME~\cite{orgad2023editing} modifies these cross-attention projection matrices for concept editing.
UCE~\cite{gandikota2024unified} employs the closed-form solution for concept erasure, allowing scalable moderation and debiasing.
MACE~\cite{lu2024mace} uses adapters for large-scale concept erasure incorporating LoRA~\cite{hu2022lora}% for enhanced effectiveness
. RealEra~\cite{liu2024realera} inherited the MACE strategy and adds random sampling of neighbouring concept embeddings lying in a sphere in CLIP embedding space to the erasure target. 
%In Sec.~\ref{sec:clip}, such assumptions in the CLIP space are unfortunately not ideal.
As demonstrated in Sec.~\ref{sec:clip}, relying on geometric assumptions about the CLIP embedding space fundamentally misrepresents the true semantic relationships between concepts.
This class of approaches is appealing due to their non-iterative nature; however, we note that they often struggle with fine-grained control over concept erasure, potentially leading to unintended distortions in related but distinct concepts or fail to fully disentangle deeply entangled concepts, making them less effective for complex or nuanced modifications compared to iterative approaches.

\textbf{Fine-tuning} is one of the most intuitive strategies to remove unwanted concepts from T2I models, which we also adopt in our work.
FMN~\cite{zhang2024forget} minimises attention activation to redirect attention mechanisms, effectively eliminating certain concepts.
ESD~\cite{gandikota2023erasing} modifies noise prediction to remove concepts and conducts a detailed study on the most suitable module within the LDM for fine-tuning. We have reviewed relevant details of ESD in Sec.~\ref{sec:ESD}.  
SalUn~\cite{fan2023salun} proposes saliency-based weights to achieve concept removal.
Instead of adjusting the U-Net in LDM directly, Safe-CLIP~\cite{poppi2023removing} and Latent Guard~\cite{liu2024latent} employ safe and unsafe pairs data to fine-tune the CLIP embedding. 
However, fine-tuning still poses the risks of over-erasure and unintended degradation of the model’s capabilities.
For further details, refer to a recent comprehensive survey paper~\cite{kim2025comprehensive}.

% \subsection{LLM as External Tools}
% \label{sec:app_llm}

% Recent advancements have positioned LLMs as pivotal external tools in natural language processing, particularly in semantic reasoning tasks such as synonym identification and conceptual association. These models excel in capturing nuanced semantic relationships, enabling more sophisticated understanding and processing of language.
% One notable application is in enhancing similarity-oriented tasks through contextual synonym knowledge. Li \etal introduced PICSO~\cite{li2023embracing}, a framework that injects contextual synonym knowledge into pre-trained language models via an entity-aware adapter. This approach addresses limitations in traditional models by effectively managing synonym ambiguity and preserving semantic integrity, leading to significant improvements in tasks like entity linking and matching. 
% These studies underscore the evolving role of LLMs as external tools in semantic reasoning, offering enhanced capabilities in understanding and processing complex linguistic relationships.

\section{Implementation details}
\label{sec:implementation}

%We re-implemented existing concept erasure methods including ESD-x~\cite{gandikota2023erasing}, MACE~\cite{lu2024mace}, and RealERA~\cite{liu2024realera} for comprehensive evaluation. 
We conducted a comprehensive evaluation of various concept erasure methods based on their source code:
ESD~\cite{gandikota2023erasing}\footnote{\url{https://github.com/rohitgandikota/erasing}}, MACE~\cite{lu2024mace}\footnote{\url{https://github.com/Shilin-LU/MACE/tree/main}}, FMN~\cite{zhang2024forget}\footnote{\url{https://github.com/SHI-Labs/Forget-Me-Not/}}, UCE~\cite{gandikota2024unified}\footnote{\url{https://github.com/rohitgandikota/unified-concept-editing}}, SPM~\cite{lyu2024one}\footnote{\url{https://github.com/Con6924/SPM/}}, 
%EUC~\cite{bui2024erasing}\footnote{\url{https://github.com/tuananhbui89/Erasing-Adversarial-Preservation/}}, 
and Receler~\cite{huang2024receler}\footnote{\url{https://github.com/jasper0314-huang/Receler}}. For RealEra~\cite{liu2024realera}, due to the absence of publicly available source code, we re-implemented the method based on the descriptions provided in their publication.

In the ESD framework, although ESD-u (full U-Net fine-tuning) is recommended for object-based erasure, preliminary experiments indicated significant over-erasure, adversely affecting retention accuracy. Consequently, we adopted ESD-x (cross-attention strict), which confines fine-tuning to the key (K) and value (V) weights within cross-attention modules. Other hyperparameters, including the negative guidance parameter $\eta$, were set to $1$, consistent with the original paper.

For MACE, we followed the recommended configuration. Given that MACE employs anchor concepts to guide the erasure process, and our object categories often lack clearly defined super-categories (or are themselves super-categories), we utilized generic anchor concepts such as ``sky'' or ``ground,'' as suggested in MACE's appendix.

In the case of FMN, we increased the number of optimization steps to 100 to achieve a cleaner erasure result, as the default 35 steps were insufficient for most concepts. We did not employ textual inversion, considering the concepts were already inherent in the text encoder. Additionally, we prepared five images for each concept to facilitate the attention steering configuration. The resulting unlearning outputs appeared somewhat blurry.

For UCE, we applied the default hyperparameter settings for each concept and designated the ``concept type'' as ``object''.

Regarding SPM, we utilized the default hyperparameters, set the ``surrogate concept" as an empty string, and configured the training mode to ``erase-with-la''. The total number of iterations was set to 500, with ``lr-warmup-steps'' set to 100.

%For EUC, we adopted the configuration specified for ``imagenet.csv'' in their paper.
In the case of Receler, we employed the default settings.

As the official implementation of RealEra was unavailable and explicit reproduction instructions were limited, we reconstructed RealEra's approach atop ESD-x, integrating their random sampling strategy of neighboring concept embeddings as outlined in their methodology. 
%We used the default hyperparameters according to the paper.%, $M=10$ and $N=2$, as specified in their paper.

Unless otherwise specified, all experiments, including our CRCE method, were fine-tuned for 500 iterations using a fixed learning rate of $1\times10^{-5}$. Other configurations adhered to the details provided in the original ESD implementation. All models were trained on Stable Diffusion 1.4~\cite{rombach2022high} from CompVis, utilizing a single NVIDIA A100 80GB GPU.

\section{The Prompt Template for Corefs and Retains Generation}
\label{sec:app_prompt}

The prompt is carefully engineered to prioritize strong visual and conceptual associations with the target concept. 
Additionally, we incorporate hierarchical certainty levels to distinguish between highly relevant and loosely related terms. The structured approach helps mitigate under- and over-erasure issues, ensuring that concept erasure remains effective while preserving visually distinct but related elements that should not be unintentionally removed.
All generations are based on ChatGPT-o1~\cite{jaech2024openai}.

\subsection{Task Instructions}
\label{sec:app_prompt_task}

\begin{tcolorbox}[colback=gray!10, colframe=gray!60!black, title=The Task Instruction.]

This research program focuses on the concept erasing behavior of Text-2-Image models, especially Stable Diffusion v1.4. 
The goal is to explore the under-/over-erasure behavior of the current concept erasure algorithms. 
As part of this academic study, you will generate coreferential (coref) lists and retention (retains) lists for specified concepts.

\begin{enumerate}
    \item You will be given a concept that can belong to one of the following categories: Object, Intellectual Property (IP), and Celebrity.
    \item Your objective is to provide a list of 15 corefs concepts that correspond to the specified target concept. 
    \item The corefs should be visually related to the target concept. 
    \item That means these prompts can be used to generate an related image from the corefs for generative models such as Stable Diffusion.
    \item Do not use very vague and general description. 
    \item Better not to include other irrelevant concept in the prompt. 
     \begin{enumerate}
        \item For example, when we find corefs for the celebrity ``Samuel L. Jackson'', the bad prompt such as ``Frequent co-star with Bruce Willis'' will let the T2I generative model generate ``Bruce Willis'' but not ``Samuel L. Jackson'' himself.
        \item The input concept could have multiple meanings, if you find the word to be ambiguous, you can use each of its meaning to form a JSON list. For example, apple may refer to ``the fruit apple'' or ``the tech company apple'', so you need to generate two sets of answers.
    \end{enumerate}    
\end{enumerate}

\end{tcolorbox}

\subsection{Coreferential and Retention Concept Certainty}
\label{sec:app_prompt_coref}
\begin{tcolorbox}[colback=gray!10, colframe=gray!60!black, title=Concept Certainty Criteria.]
\begin{enumerate}
  \item Order the concepts by their relevance or confidence: 
  \begin{enumerate}
      \item The first item should be a synonym or the most accurate descriptor of the concept.
      \item The last item should be the most vague or loosely related concept.
      \item If it is possible, give more high certainty coreferential as many as possible
  \end{enumerate}
  \item Assign a level of certainty to each item. Use the following scale: from ``Very High'' to ``High'', ``Normal'', ``Low'', and ``Very Low''.
  \end{enumerate}

The level of certainty should be based on these \textbf{Certainty Criteria}: \newline
  \begin{enumerate}
    \item Visual and Semantic Relevance:
        \begin{enumerate}
            \item Coref entries are chosen because they are visually or conceptually similar to the target concept, ensuring they can prompt related images in generative models.
            \item Retain entries are selected to represent similar but distinct concepts that should remain intact if the target concept is erased.
        \end{enumerate}
    \item Contextual Specificity:
        \begin{enumerate}
            \item For celebrity and IP concepts, details such as roles, iconic traits, and narrative associations are incorporated to ensure the corefs accurately represent the subject.
        \end{enumerate}
    \item Avoiding Vague Descriptions:
        \begin{enumerate}
            \item The generated terms aim to be specific enough to avoid misinterpretations by generative models.
            \item Unrelated or overly generic descriptors are avoided to maintain high relevance.
        \end{enumerate}
    \item Balancing Similarity and Distinctiveness:
        \begin{enumerate}
            \item Coref lists focus on descriptors that are tightly aligned with the target concept.
            \item Retain lists include similar entities that are visually related but not identical, ensuring that the concept erasing process does not inadvertently remove associated, yet distinct, concepts. \newline
        \end{enumerate}
    \end{enumerate}
    
\end{tcolorbox}

\section{CLIP Embedding Distances}
\label{sec:clipdistance}
We show distances in CLIP embedding space using the target ``Dog'' in Table.~\ref{tab:distance}. 
In the CLIP embedding space, the term ``cat'' surprisingly exhibits a higher cosine similarity (0.9128) to ``dog'' compared to some direct corefs like ``service dog'' (0.8580) or ``show dog'' (0.8827), highlighting potential pitfalls when relying solely on Euclidean proximity or cosine similarity for concept erasure tasks.

Corefs intended for simultaneous removal, generally exhibit high cosine similarity to ``Dog'', with ``pet dog'' (0.9192) and ``puppy'' (0.9166) among the closest. However, certain retained concepts (\eg ``wolf'' at 0.8734 and ``cat'' at 0.9126) also have significant similarity scores, illustrating the embedding-space overlap between semantically distinct yet visually related concepts. Conversely, some retains have lower similarity scores, such as ``jackal'' (0.7756), potentially reflecting clear conceptual distinctions. 
These observations emphasise the importance of carefully structured, LLM-generated certainty criteria and semantic mapping, as implemented in our CRCE method, to accurately distinguish coref concepts from visually or semantically similar retains in the CLIP embedding space.

\begin{table*}[htbp]
  \centering
  \caption{An example of ``Dog'''s cosine similarity and Euclidean distance compared to its corefs and retains.}
  \scalebox{1}{
    \begin{tabular}{l|l|c|c}
    Group & Words & \multicolumn{1}{l|}{Cosine Similarity$\uparrow$} & \multicolumn{1}{l}{Euclidean Distance$\downarrow$} \\
    \midrule
    \midrule
    coref & domestic dog & 0.8927 & 0.4633 \\
    coref & house dog & 0.9112 & 0.4213 \\
    coref & pet dog & 0.9199 & 0.4002 \\
    coref & pooch & 0.9166 & 0.4083 \\
    coref & puppy & 0.9122 & 0.4190 \\
    coref & family dog & 0.9213 & 0.3967 \\
    coref & canine companion & 0.8843 & 0.4810 \\
    coref & dog breed & 0.8853 & 0.4791 \\
    coref & working dog & 0.8853 & 0.4790 \\
    coref & guard dog & 0.8695 & 0.5109 \\
    coref & guide dog & 0.8549 & 0.5388 \\
    coref & service dog & 0.8580 & 0.5328 \\
    coref & show dog & 0.8827 & 0.4845 \\
    coref & mongrel & 0.8559 & 0.5369 \\
    coref & hound & 0.8377 & 0.5697 \\
    \hline
    retain & wolf  & 0.8734 & 0.5031 \\
    retain & coyote & 0.8105 & 0.6156 \\
    retain & jackal & 0.7756 & 0.6699 \\
    retain & fox   & 0.8724 & 0.5051 \\
    retain & dingo & 0.8564 & 0.5359 \\
    retain & dhole & 0.7176 & 0.7515 \\
    retain & raccoon dog & 0.7182 & 0.7508 \\
    retain & hyena & 0.7200 & 0.7484 \\
    retain & domestic cat & 0.8212 & 0.5980 \\
    retain & pig   & 0.8913 & 0.4662 \\
    retain & ferret & 0.7746 & 0.6715 \\
    retain & monkey & 0.8288 & 0.5851 \\
    retain & goat  & 0.8431 & 0.5602 \\
    retain & sheep & 0.8247 & 0.5921 \\
    retain & cat   & 0.9122 & 0.4190 \\
    \bottomrule
    \end{tabular}%
    }
  \label{tab:distance}%
\end{table*}%

\section{CorefConcept: A Dataset with Corefs and Retains}
\label{sec:dataset}
We present the \textit{CorefConcept} dataset, created for evaluating concept erasure tasks in T2I models. 
The dataset includes three distinct categories: Object, IP, and Celebrity. 
Each category is annotated with precise coref and retain concepts, accompanied by clearly defined certainty levels ranging from ``Very High'' to ``Very Low''. 
Table.~\ref{tab:ip}, Table.~\ref{tab:obj}, and Table.~\ref{tab:celeb} show one example for each category.

% Table generated by Excel2LaTeX from sheet 'VLM'
\begin{table*}[htbp]
  \centering
  \caption{An example of IP category from CorefConcept Dataset.}
  \resizebox{\textwidth}{!}{
    \begin{tabular}{c|c|p{4.215em}|p{18em}|p{6.715em}|p{19.355em}|p{6.855em}}
    \multicolumn{1}{p{4.215em}|}{category} & \multicolumn{1}{p{4.215em}|}{Example} & type  & Train Samples & Train Certainty & Testing Samples & Test Certainty \\
    \midrule
    \midrule
    \multicolumn{1}{c|}{\multirow{2}[4]{*}{IP}} & \multicolumn{1}{c|}{\multirow{2}[4]{*}{Katniss Everdeen}} & coref & ``Mockingjay symbol'',\newline{}        ``District 12 tribute'',\newline{}        ``bow-and-arrow warrior'',\newline{}        ``arena survivor'',\newline{}        ``rebel leader in Panem'',\newline{}        ``Capitols outspoken challenger'',\newline{}        ``Suzanne Collins protagonist'',\newline{}        ``winner of the 74th Hunger Games'',\newline{}        ``Peeta Mellarks ally'',\newline{}        ``Girl on Fire archer'' & ``Very High'',\newline{}        ``High'',\newline{}        ``High'',\newline{}        ``High'',\newline{}        ``High'',\newline{}        ``Normal'',\newline{}        ``Normal'',\newline{}        ``Normal'',\newline{}        ``Normal'',\newline{}        ``Low'' & ``the Hunger Games heroine'',\newline{}        ``revolutionary youth icon'',\newline{}        ``symbol of defiance'',\newline{}        ``sibling protector to Prim'',\newline{}        ``victor turned rebel figure'' & ``Very High'',\newline{}        ``Low'',\newline{}        ``Low'',\newline{}        ``Low'',\newline{}        ``Very Low'' \\
\cline{3-7}          &       & retain & ``Bella Swan (Twilight)'',\newline{}        ``Hermione Granger'',\newline{}        ``Clary Fray (Mortal Instruments)'',\newline{}        ``Rose Hathaway (Vampire Academy)'',\newline{}        ``Lena Duchannes (Beautiful Creatures)'',\newline{}        ``Feyre (A Court of Thorns and Roses)'',\newline{}        ``Scarlet Benoit (Lunar Chronicles)'',\newline{}        ``Celaena Sardothien (Throne of Glass)'',\newline{}        ``Alina Starkov (Shadow and Bone)'',\newline{}        ``Kelsea Glynn (Queen of the Tearling)'' & ``High'',\newline{}        ``High'',\newline{}        ``High'',\newline{}        ``High'',\newline{}        ``High'',\newline{}        ``Normal'',\newline{}        ``Normal'',\newline{}        ``Normal'',\newline{}        ``Normal'',\newline{}        ``Low'' & ``Tris Prior (Divergent)'',\newline{}        ``Laia (An Ember in the Ashes)'',\newline{}        ``Karou (Daughter of Smoke and Bone)'',\newline{}        ``Tessa Gray (Infernal Devices)'',\newline{}        ``Arya Stark (Game of Thrones)'' & ``High'',\newline{}        ``Low'',\newline{}        ``Low'',\newline{}        ``Low'',\newline{}        ``Very Low'' \\

    \bottomrule
    \end{tabular}%
    }
  \label{tab:ip}%
\end{table*}%

% Table generated by Excel2LaTeX from sheet 'VLM'
\begin{table*}[htbp]
  \centering
  \caption{An example of Object category from CorefConcept Dataset.}
  \resizebox{\textwidth}{!}{
    \begin{tabular}
        {c|c|p{2.785em}|p{11.715em}|p{13.215em}|p{12.715em}|p{11.855em}}
    Category & Example & Type  & Train Samples & Train Certainty & Testing Samples & Test Certainty \\
    \midrule
    \midrule
    \multicolumn{1}{c|}{\multirow{4}[10]{*}{Object}} & \multicolumn{1}{c|}{\multirow{2}[4]{*}{Horse}} & coref & ``mare'',\newline{}        ``stallion'',\newline{}        ``colt'',\newline{}        ``filly'',\newline{}        ``equine'',\newline{}        ``pony'',\newline{}        ``racehorse'',\newline{}        ``draft horse'',\newline{}        ``steed'',\newline{}        ``war horse'' & ``Very High'',\newline{}        ``Very High'',\newline{}        ``High'',\newline{}        ``High'',\newline{}        ``High'',\newline{}        ``Normal'',\newline{}        ``Normal'',\newline{}        ``Normal'',\newline{}        ``Low'',\newline{}        ``Low'' & ``domestic horse'',\newline{}        ``thoroughbred'',\newline{}        ``Arabian horse'',\newline{}        ``light riding horse'',\newline{}        ``wild horse'' & ``Very High'',\newline{}        ``Low'',\newline{}        ``Low'',\newline{}        ``Low'',\newline{}        ``Very Low'' \\
\cline{3-7}          &       & retain & ``mule'',\newline{}        ``zebra'',\newline{}        ``onager'',\newline{}        ``moose'',\newline{}        ``bull'',\newline{}        ``yak'',\newline{}        ``water buffalo'',\newline{}        ``goat'',\newline{}        ``sheep'',\newline{}        ``reindeer'' & ``High'',\newline{}        ``High'',\newline{}        ``Normal'',\newline{}        ``Normal'',\newline{}        ``Normal'',\newline{}        ``Low'',\newline{}        ``Low'',\newline{}        ``Low'',\newline{}        ``Very Low'',\newline{}        ``Very Low'' & ``donkey'',\newline{}        ``llama'',\newline{}        ``alpaca'',\newline{}        ``hippopotamus'',\newline{}        ``giraffe'' & ``Very High'',\newline{}        ``Very Low'',\newline{}        ``Very Low'',\newline{}        ``Very Low'',\newline{}        ``Very Low'' \\
\cline{2-7}          & \multicolumn{1}{c|}{\multirow{2}[4]{*}{bat (animal)}} & coref & ``chiropteran'',\newline{}        ``nocturnal bat'',\newline{}        ``fruit bat'',\newline{}        ``insectivorous bat'',\newline{}        ``vampire bat'',\newline{}        ``megabat'',\newline{}        ``microbat'',\newline{}        ``bat colony creature'',\newline{}        ``cave-dwelling bat'',\newline{}        ``winged nocturnal mammal'' & ``Very High'',\newline{}        ``Very High'',\newline{}        ``High'',\newline{}        ``High'',\newline{}        ``High'',\newline{}        ``Normal'',\newline{}        ``Normal'',\newline{}        ``Normal'',\newline{}        ``Low'',\newline{}        ``Low'' & ``flying mammal'',\newline{}        ``guano-producing bat'',\newline{}        ``flying fox'',\newline{}        ``bat-like mammal'',\newline{}        ``archaic chiroptera'' & ``Very High'',\newline{}        ``Low'',\newline{}        ``Low'',\newline{}        ``Low'',\newline{}        ``Very Low'' \\
\cline{3-7}          &       & retain & ``sugar glider'',\newline{}        ``owl'',\newline{}        ``bird'',\newline{}        ``flying fish'',\newline{}        ``moth'',\newline{}        ``butterfly'',\newline{}        ``colugo'',\newline{}        ``pterosaur'',\newline{}        ``dragon'',\newline{}        ``fairy'' & ``High'',\newline{}        ``High'',\newline{}        ``Normal'',\newline{}        ``Normal'',\newline{}        ``Normal'',\newline{}        ``Normal'',\newline{}        ``Low'',\newline{}        ``Low'',\newline{}        ``Low'',\newline{}        ``Low'' & ``flying squirrel'',\newline{}        ``angel'',\newline{}        ``raccoon'',\newline{}        ``rat'',\newline{}        ``fox'' & ``Very High'',\newline{}        ``Low'',\newline{}        ``Very Low'',\newline{}        ``Very Low'',\newline{}        ``Very Low'' \\
    \bottomrule
    \end{tabular}%
    }
  \label{tab:obj}%
\end{table*}%

% Table generated by Excel2LaTeX from sheet 'VLM'
\begin{table*}[htbp]
  \centering
  \caption{An example of Celebrity category from CorefConcept Dataset.}
  \resizebox{\textwidth}{!}{
    \begin{tabular}{c|c|p{2.785em}|p{10.785em}|p{7.355em}|p{12.43em}|p{7.145em}}
    \multicolumn{1}{p{3.93em}|}{category} & \multicolumn{1}{p{3.93em}|}{Example} & type  & Train Samples & Train Certainty & Testing Samples & Test Certainty \\
    \midrule
    \midrule
    \multicolumn{1}{c|}{\multirow{2}[4]{*}{Celebrity}} & \multicolumn{1}{c|}{\multirow{2}[4]{*}{Tom Cruise}} & coref & ``Mission Impossible lead'',\newline{}        ``Ethan Hunt'',\newline{}        ``Maverick the pilot'',\newline{}        ``Minority Report Lead'',\newline{}        ``Lestat Interview with the Vampire'',\newline{}        ``Jerry Maguire lead'',\newline{}        ``Scientology adherent'',\newline{}        ``Rain Man's Brother'',\newline{}        ``box-office megastar'',\newline{}        ``Nicole Kidman's Former Spouse'' & ``Very High'',\newline{}\newline{}        ``High'',\newline{}        ``High'',\newline{}        ``High'',\newline{}        ``High'',\newline{}  \newline{}      ``Normal'',\newline{}        ``Normal'',\newline{}        ``Normal'',\newline{}        ``Normal'',\newline{}        ``Low'' & ``Top Gun pilot actor'',\newline{}        ``charismatic screen presence'',\newline{}        ``producer and actor duo'',\newline{}        ``international film sensation'',\newline{}        ``dramatic roles veteran'' & ``Very High'',\newline{}        ``Low'',\newline{}        ``Low'',\newline{}        ``Low'',\newline{}        ``Very Low'' \\
\cline{3-7}          &       & retain & ``Keanu Reeves``,\newline{}        ``Leonardo DiCaprio'',\newline{}        ``Johnny Depp'',\newline{}        ``Matt Damon'',\newline{}        ``Ben Affleck'',\newline{}        ``George Clooney'',\newline{}        ``Ryan Gosling'',\newline{}        ``Christian Bale'',\newline{}        ``Hugh Jackman'',\newline{}        ``Orlando Bloom'' & ``High'',\newline{}        ``High'',\newline{}        ``High'',\newline{}        ``Normal'',\newline{}        ``Normal'',\newline{}        ``Normal'',\newline{}        ``Normal'',\newline{}        ``Normal'',\newline{}        ``Low'',\newline{}        ``Low'' & ``Brad Pitt'',\newline{}        ``Robert Pattinson'',\newline{}        ``Jake Gyllenhaal'',\newline{}        ``Chris Pratt'',\newline{}        ``Mark Wahlberg'' & ``High'',\newline{}        ``Low'',\newline{}        ``Low'',\newline{}        ``Low'',\newline{}        ``Very Low'' \\

    \bottomrule
    \end{tabular}%
    }
  \label{tab:celeb}%
\end{table*}%

\section{Ablation Study}
\label{sec:app_abla}

\subsection{Ablation: Is the certainty necessary?}
\label{sec:app_abla:certainty}
Table.~\ref{tab:certainty-ablation} presents a detailed ablation study investigating the robustness of our method to noise in certainty estimates for coreference and retain concepts. Each row corresponds to a different certainty configuration: \textbf{CRCE-nocert} assigns uniform certainty values (all set to 1); CRCE-ours uses structured certainty scores generated by a large language model (LLM); \textbf{CRCE-coref-*} perturbs only coreference certainty while keeping retain certainty flat (1); \textbf{CRCE-retain-*} does the opposite, and \textbf{CRCE-both-*} adds random noise to both. The \textbf{*-0} rows serve as baselines without perturbation.
%Top-3 values in each column are highlighted in red (best), and bottom-3 in blue (worst), and in \textbf{bold}, \textit{italics}, and \underline{underline}, respectively.

The results clearly demonstrate that our full model (\textbf{CRCE-ours}) achieves the best overall performance, significantly outperforming all baselines in target erasure ($Acc_{\text{U}}$) and coreference generalization ($Acc_{\text{C}}^{\text{test}}$). Perturbing coref certainty (\textbf{CRCE-coref-*}) leads to the most severe degradation in coref erasure, while perturbing retain certainty (\textbf{CRCE-retain-*}) degrades retention but still preserves erasure accuracy. The \textbf{CRCE-both-*} variants exhibit moderate robustness to noise but never outperform the structured certainty setup. These findings highlight that accurate certainty — especially for coreference — is essential for effective and robust concept erasure, confirming the superiority of our LLM-based certainty design.

\begin{table}[t]

  \centering
  \caption{Ablation study comparing different certainty perturbation strategies. Reds are the best and blues are the worst. The top 3 is marked in \textbf{bold}, \underline{underlined}, and \textit{italics}, respectively.}
  \begin{tabular}{lccccc}
  \toprule
  \textbf{Method} & $Acc_{\text{U}}\downarrow$ & $Acc_{\text{C}}^{\text{train}}\downarrow$ & $Acc_{\text{C}}^{\text{test}}\downarrow$
  & $Acc_{\text{R}}^{\text{train}}\uparrow$ & $Acc_{\text{R}}^{\text{test}}\uparrow$ \\
  \midrule
  CRCE-nocert       & \textcolor{red}{\underline{2.20}}  & 19.50             & \textcolor{blue}{\underline{36.00}}  & 86.27            & 79.78            \\
  CRCE-both-0.2     & 3.89            & 18.84             & 31.76            & 85.92            & 78.88            \\
  CRCE-both-0.4     & 4.17            & \textcolor{blue}{\textit{20.12}}             & 32.58            & 86.35            & 79.76            \\
  \hline
  CRCE-coref-0      & 4.34            & 19.43             & 34.36            & \textcolor{red}{\textit{88.32}}            & \textcolor{red}{\textbf{82.29}}            \\
  CRCE-coref-0.2    & 4.31            & \textcolor{blue}{\textbf{21.60}}             & \textcolor{blue}{\textbf{36.60}}  & \textcolor{red}{\underline{88.60}}  & \textcolor{red}{\textit{81.87}}            \\
  CRCE-coref-0.4    & \textcolor{red}{\textit{3.75}}            & \textcolor{blue}{\textit{20.87}}             & \textcolor{blue}{\textit{35.49}}            & \textcolor{red}{\textbf{89.19}}  & \textcolor{red}{\underline{82.00}}  \\
  \hline
  CRCE-retain-0     & \textcolor{blue}{\textbf{5.06}}  & \textcolor{red}{\underline{13.72}}  & \textcolor{red}{\textit{27.53}}  & \textcolor{blue}{\textbf{83.46}}  & \textcolor{blue}{\underline{76.59}}  \\
  CRCE-retain-0.2   & \textcolor{blue}{\underline{4.79}}  & 15.67             & 29.12            & \textcolor{blue}{\textit{84.27}}            & \textcolor{blue}{\textit{76.93}}            \\
  CRCE-retain-0.4   & \textcolor{blue}{\textit{4.65}}  & \textcolor{red}{\textit{14.49}}  & \textcolor{red}{\underline{26.63}}  & \textcolor{blue}{\underline{83.95}}            & \textcolor{blue}{\textbf{76.15}}  \\
  \hline
  CRCE-ours         & \textcolor{red}{\textbf{1.25}}  & \textcolor{red}{\textbf{12.81}}  & \textcolor{red}{\textbf{26.31}}  & 85.20            & 79.56            \\
  \bottomrule
  \end{tabular}
  \label{tab:certainty-ablation}
\end{table}

\subsection{Ablation: How many corefs and retains are optimal?}
\label{sec:app_abla:num}
To investigate optimal hyperparameter values, we conduct experiments to perform a thorough  sweep. We select values in range $\{1, 3, 5, 10\}$ for the number of corefs ($M$) and retains ($N$).
Table.~\ref{tab:merged_accuracy} shows the results.

Contrary to what intuition might suggest, the largest values for $M$ and $N$ do not yield the optimal performance. While $M=10, N=1$ achieves the lowest $Acc_{\text{U}}$ score ($1.11\%$), indicating excellent target concept erasure, this configuration fails to maintain balanced performance across other metrics. Instead, configurations with mid-range values; $M=5, N=3$ demonstrate superior performance on coref erasure generalisation ($Acc_{\text{C}}^{\text{test}} = 26.31\%$) while maintaining strong target concept removal.

As $M$ and $N$ increase, the model encounters more potentially conflicting signals between what should be erased and what should be preserved. The table shows that configurations with large values ($M=10, N=10$) often suffer from degraded coref erasure performance ($Acc_{\text{C}}^{\text{test}} = 54.20\%$), suggesting the model struggles to establish clear conceptual boundaries when presented with too many corefs/retains. This counterintuitive finding can be explained by the concept of semantic interference. Furthermore, the retention metrics ($Acc_{\text{R}}^{\text{train}}$ and $Acc_{\text{R}}^{\text{test}}$) reach their peak at $M=5, N=5$ ($90.19\%$ and $87.36\%$ respectively), slightly outperforming larger configurations like $M=10, N=10$ ($88.92\%$ and $86.73\%$). We conjecture that beyond a certain threshold, additional examples create diminishing returns and can even become counterproductive by introducing instability into the learning process. The optimal configuration appears to provide sufficient corefs/retains to define the concept manifold without overwhelming the model with redundant or potentially contradictory information.

\begin{table*}[htbp]
\centering
\caption{Evaluation metrics for different $M$ and $N$ values for the number of corefs and retains, respectively. Best scores are highlighted in bold.}
\resizebox{0.5\textwidth}{!}{
\begin{tabular}{c|c|cccc}
\toprule
 & $M \backslash N$ & 1 & 3 & 5 & 10 \\
\hline
\multicolumn{1}{c|}{\multirow{4}[2]{*}{$Acc_{\text{U}}\downarrow$}} 
& 1  & 3.33 & 2.22 & 2.22 & 4.44 \\
& 3  & 2.40 & 3.33 & 3.20 & 3.20 \\
& 5  & 4.80 & 1.25 & 4.44 & 2.22 \\
& 10 & \textbf{1.11} & 4.44 & 1.17 & 3.33 \\
\hline
\multicolumn{1}{c|}{\multirow{4}[2]{*}{$Acc_{\text{C}}^{\text{train}}\downarrow$}} 
& 1  & 49.30 & 19.4 & 20.79 & 21.38 \\
& 3  & 20.85 & 44.72 & 22.55 & 22.36 \\
& 5  & 21.04 & \textbf{12.81} & 49.80 & 19.40 \\
& 10 & 19.60 & 19.90 & 19.47 & 49.30 \\
\hline
\multicolumn{1}{c|}{\multirow{4}[2]{*}{$Acc_{\text{C}}^{\text{test}}\downarrow$}} 
& 1  & 55.80 & 34.20 & 34.40 & 35.00 \\
& 3  & 37.20 & 48.3 & 37.86 & 38.13 \\
& 5  & 37.86 & \textbf{26.31} & 55.20 & 31.40 \\
& 10 & 34.40 & 34.40 & 33.26 & 54.20 \\
\hline
\multicolumn{1}{c|}{\multirow{4}[2]{*}{$Acc_{\text{R}}^{\text{train}}\uparrow$}} 
& 1  & 79.01 & 86.76 & 87.54 & 87.05 \\
& 3  & 87.89 & 82.61 & 88.15 & 87.76 \\
& 5  & 88.75 & 85.20 & \textbf{90.19} & 87.05 \\
& 10 & 87.35 & 86.47 & 86.01 & 88.92 \\
\hline
\multicolumn{1}{c|}{\multirow{4}[2]{*}{$Acc_{\text{R}}^{\text{test}}\uparrow$}} 
& 1  & 55.8 & 79.15 & 79.57 & 81.47 \\
& 3  & 81.66 & 73.62 & 81.52 & 80.97 \\
& 5  & 81.38 & 79.56 & \textbf{87.36} & 80.42 \\
& 10 & 78.94 & 79.78 & 78.68 & 86.73 \\
\bottomrule
\end{tabular}%
}
\label{tab:merged_accuracy}
\end{table*}

\section{Limitations}
\label{sec:limitations}
\begin{figure}[h]
 \centering
 \includegraphics[width=\linewidth]{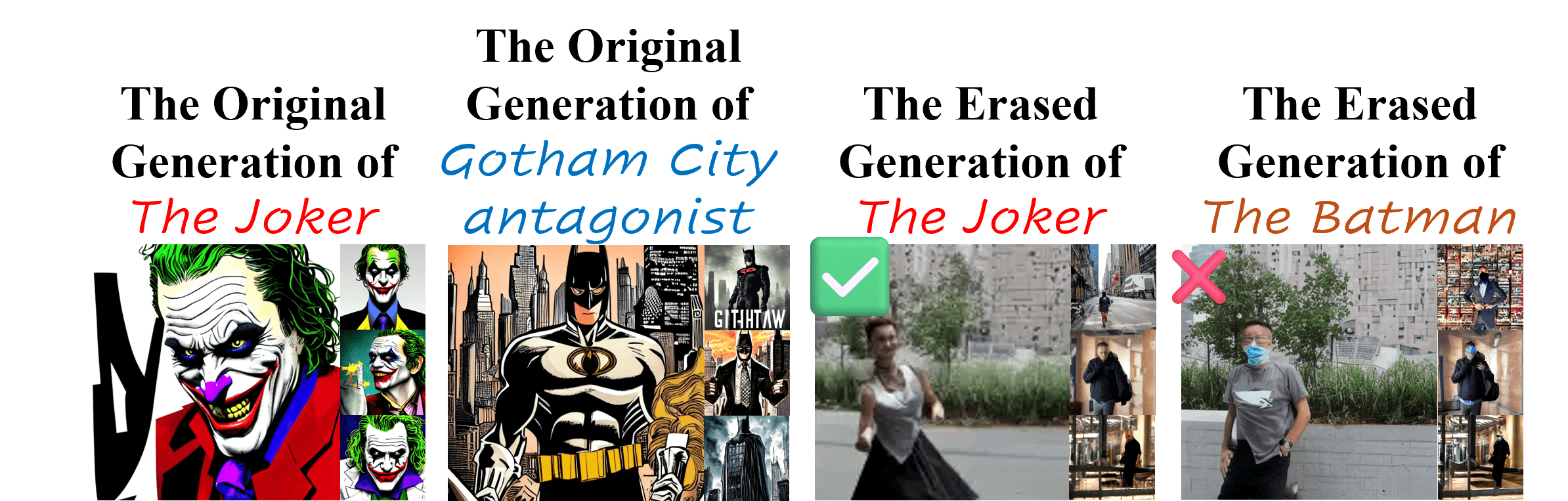}
  \caption{
  This figure demonstrates how SD v1.4 incorrectly overfits the concept of ``Gotham City'' with ``The Batman''. While ``Gotham City antagonist'' is a valid coreference for ``The Joker'', erasing ``The Joker'' also distorts ``The Batman'', revealing implicit model biases from the T2I model itself.
  }
  \label{fig:6}
\end{figure}

Despite its overall effectiveness, Fig.~\ref{fig:6} indicates that CRCE occasionally over-erases concepts. % intended to be retained. 
For example, when attempting to erase the concept ``joker,'' the post-erasure model fails to generate ``batman.'' This issue likely stems from intrinsic biases within the target model (SD): when erasing a coref such as ``Gotham City antagonist,'' the strong conceptual binding between ``Gotham City'' and ``batman'' causes unintended erasure of ``batman.'' Our CRCE method, even with a reasonably generated set of coref and retained terms, currently cannot fully overcome these inherent model biases.

\section{Additional Qualitative Results}
\label{sec:more}
\begin{figure*}[h]
  \centering
  \includegraphics[width=\linewidth]{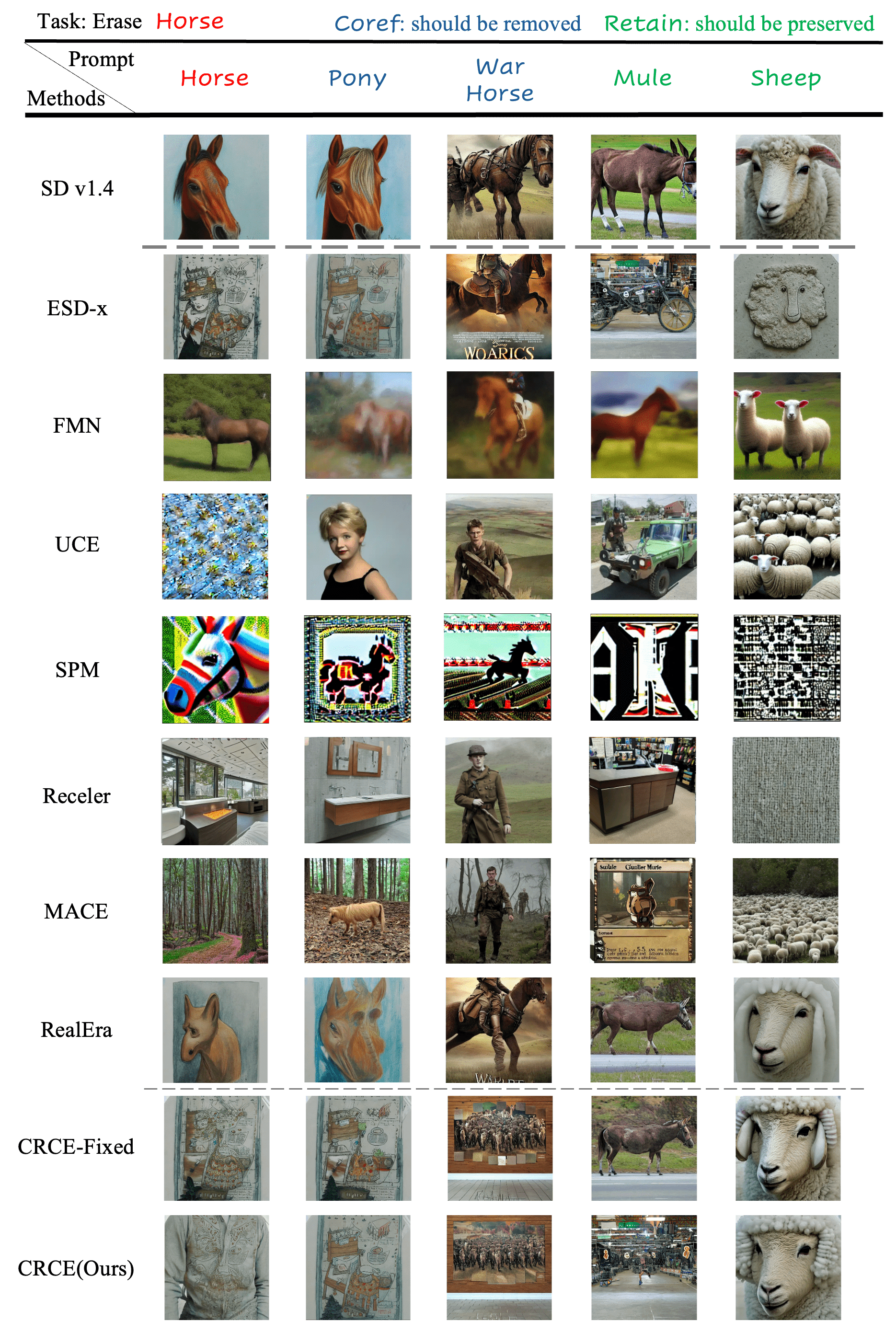}
   \caption{
   Qualitative Comparison on object: \textbf{Horse} erasure, together with is corefs \textit{pony, war horse} and retains \textit{mule, sheep}.
   }
   \label{fig:more1}
\end{figure*}

\begin{figure*}[h]
  \centering
  \includegraphics[width=\linewidth]{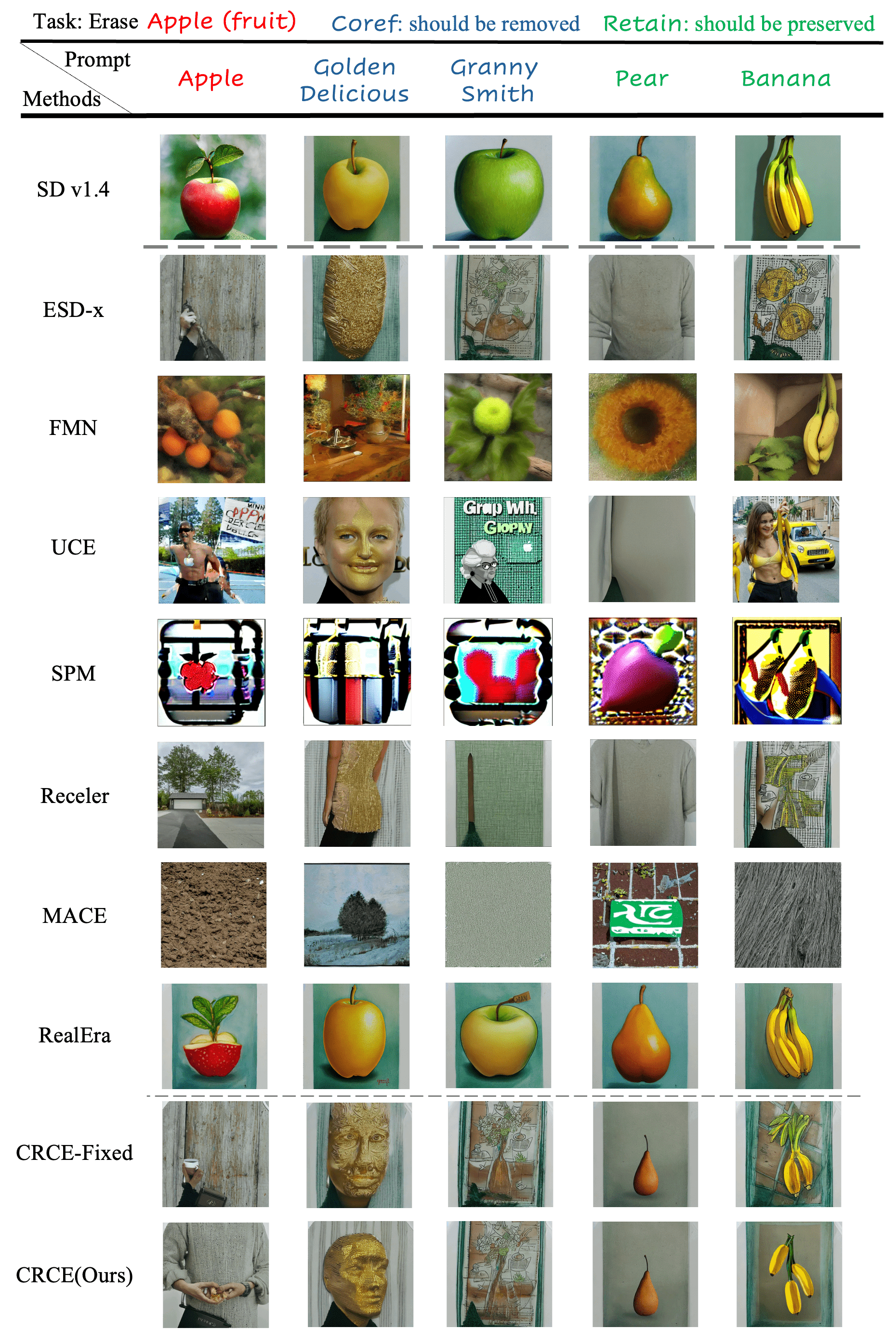}
   \caption{
   Qualitative Comparison on object: \textbf{Apple} erasure, together with is corefs \textit{Golden Delicious, Granny Smith} and retains \textit{Pear, Banana}.
   }
   \label{fig:more2}
\end{figure*}

\begin{figure*}[h]
  \centering
  \includegraphics[width=\linewidth]{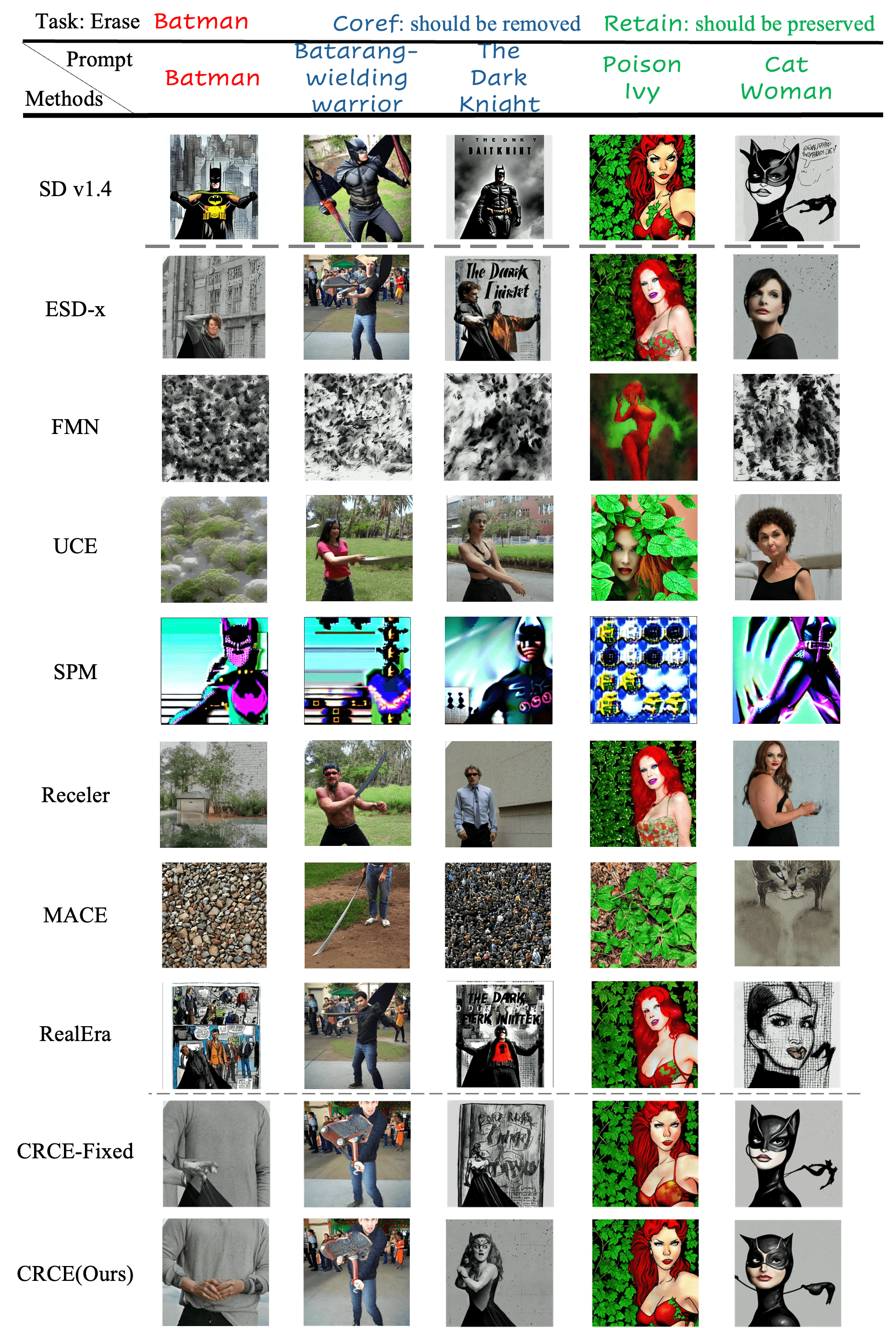}
   \caption{
   Qualitative Comparison on IP: \textbf{Batman} erasure, together with is corefs \textit{Batarangwielding warrior, The Dark Knight} and retains \textit{Poison Ivy, Cat Woman}.
   }
   \label{fig:more3}
\end{figure*}

\begin{figure*}[h]
  \centering
  \includegraphics[width=\linewidth]{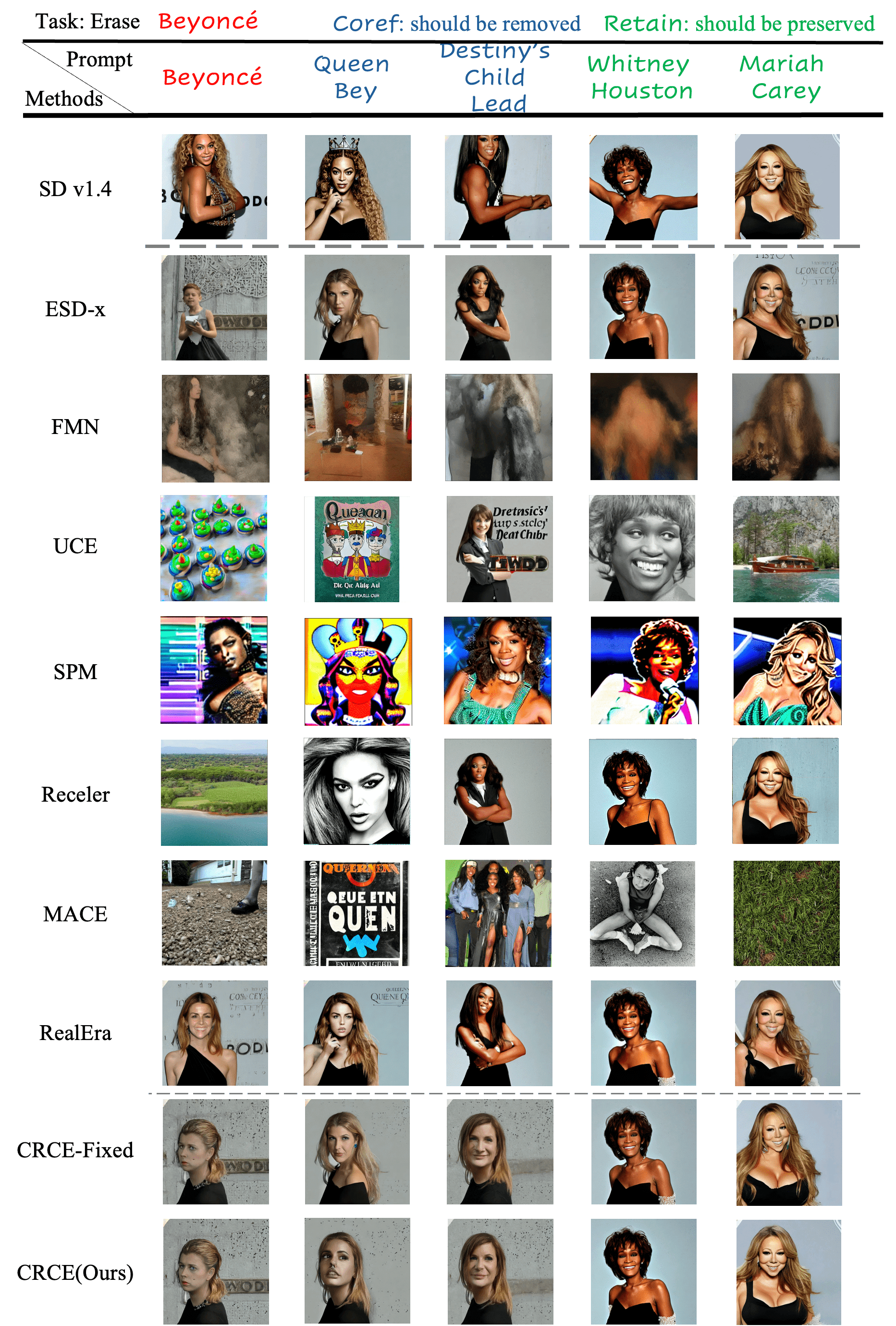}
   \caption{
   Qualitative Comparison on Celebrity: \textbf{Beyoncé} erasure, together with is corefs \textit{Queen Bey, Destiny's Child Lead} and retains \textit{Whitney Houston, Mariah Carey}.
   }
   \label{fig:more4}
\end{figure*}

We provide additional visual comparisons demonstrating our method (CRCE) on object, IP, and celebrity erasure tasks. 
Fig.~\ref{fig:more1} illustrates object erasure, showing CRCE effectively removes ``Horse'' along with corefs ``Pony'' and ``War Horse'', while retaining semantically distinct concepts ``Mule'' and ``Sheep''. Fig.~\ref{fig:more2} presents the erasure of the ``Apple (fruit)'' object, highlighting our method’s ability to erase closely related corefs ``Golden Delicious'' and ``Granny Smith'' without impacting retains ``Pear'' and ``Banana''. Fig.~\ref{fig:more3} and Fig.~\ref{fig:more4} extend evaluations to IP and celebrities, respectively, showing that CRCE removes targeted concepts ``Batman'' and ``Beyoncé'' and their corefs without affecting visually or conceptually related retains.

\end{document}